# Enhancing failure prediction in nuclear industry: Hybridization of knowledge- and data-driven techniques ☆

Amaratou Mahamadou Saley [a,b] ⓘ,*, Thierry Moyaux [b], Aïcha Sekhari [b], Vincent Cheutet [b] ⓘ, Jean-Baptiste Danielou [a]

[a] INEO Nucléaire, Lyon, France
[b] INSA Lyon, Université Lumière Lyon2 Université Claude Bernard Lyon 1, Université Jean Monnet Saint-Etienne, DISP UR4570, Villeurbanne, 69621, France

## ARTICLE INFO

## ABSTRACT

The convergence of the Internet of Things (IoT) and Industry 4.0 has significantly enhanced data-driven methodologies within the nuclear industry, notably enhancing safety and economic efficiency. This advancement challenges the precise prediction of future maintenance needs for assets, which is crucial for reducing downtime and operational costs. However, the effectiveness of data-driven methodologies in the nuclear sector requires extensive domain knowledge due to the complexity of the systems involved. Thus, this paper proposes a novel predictive maintenance methodology that combines data-driven techniques with domain knowledge from a nuclear equipment. The methodological originality of this paper is located on two levels: highlighting the limitations of purely data-driven approaches and demonstrating the importance of knowledge in enhancing the performance of the predictive models. The applicative novelty of this work lies in its use within a domain such as a nuclear industry, which is highly restricted and ultrasensitive due to security, economic and environmental concerns. A detailed real-world case study compares the current state of equipment monitoring with two scenarios, demonstrate that the methodology significantly outperforms purely data-driven methods in failure prediction. While purely data-driven methods achieve only a modest performance with a prediction horizon limited to 3 h and a F1 score of 56.36%, the hybrid approach increases the prediction horizon to 24 h and achieves a higher F1 score of 93.12%.

## 1. Introduction

In today's digital era, the nuclear industry is undergoing a significant transformation. Industry 4.0 introduces advanced technologies such as IoT and Artificial Intelligence (AI), promising unprecedented levels of operational efficiency and safety. These advancements are driving the adoption of Prognostics and Health Management (PHM) frameworks, which aim to monitor, assess, and predict the health status of critical equipment (Huang et al., 2024). PHM is widely recognized as a foundation for implementing Predictive Maintenance (PdM), by enabling the evaluation of system condition and the estimation of Remaining Useful Life (RUL) (ISO 13374-2, 2007; Peng et al., 2010). This shift toward PdM is particularly pivotal in the nuclear sector, where operational reliability and safety are paramount (Saley et al., 2022). By adopting this approach, nuclear facilities can potentially avoid unexpected shutdowns and extend the lifespan of essential equipments. Moreover, PdM enhances operational resilience while ensuring compliance with the strict safety and regulatory standards required in the industry (Saley et al., 2022).

Consequently, the use of data-driven approaches in PdM is becoming increasingly popular. The literature on this topic illustrates a variety of methods, ranging from statistical (Si et al., 2011) and probabilistic models (Özgür-Unlüakın & Bilgiç, 2006) to machine learning (ML) models (Carvalho et al., 2019). These methods analyze large amounts of data, from sensor measurements and fault histories to historical maintenance records, to detect and predict equipment failures early. ML is particularly notable for its effectiveness in managing complex and multidimensional data sets, proving vital for advanced maintenance strategies. The ML-based methods excel by learning from data, automatically adjusting their parameters, and continuously improving predictions with new data (Susto et al., 2014).

However, in the nuclear industry, the reliance on purely data-driven approaches has limitations. The complexity of nuclear equipments and the critical nature of their operational safety require more than just large data sets for accurate fault detection and prediction. Expert knowledge and deep understanding of nuclear equipment behaviors are essential to effectively interpret data and guide the learning process (Haïk et al., 2002). This underscores the need for hybrid approaches that combine ML with extensive knowledge, ensuring that predictions are not only data-driven but also informed by a comprehensive understanding of nuclear operations. This paper focuses on the prediction component of the PHM by investigating this research question: how can in-depth explicit knowledge (from documents) and tacit knowledge (from equipment experts), encompassing a comprehensive understanding of equipment monitoring, failure modes, nominal functioning, operational anomalies, and other relevant aspects, be effectively integrated across the entire failure predictive modeling workflow, from data preprocessing to model development, evaluation, and interpretation? Furthermore, this paper explores how this integration can enhance the predictive performance and operational reliability of failure prediction systems in a nuclear environment.

In this context, the main contribution of this paper is a PdM methodology for failure prediction that combines the potential of data-driven methods within the context of Industry 4.0 with enriched knowledge drawn from the operational experience on a nuclear equipment. Details on the preprocessing of this knowledge and its integration into the data preparation and failure prediction stages of the methodology are given. This approach aims to enhance the precision of predictive models, thereby contributing to the safety, efficiency, and economic viability of nuclear operations.

The remainder of this paper is organized as follows: Section 2 provides an overview of work related to data-driven methods for PdM, highlighting their limitations and emphasizing the importance of integrating knowledge to improve these approaches. Section 3 details the proposed methodology, its components, and operational logic. Section 4 presents a comprehensive case study that showcases the application of the methodology across different scenarios, highlighting its efficacy and potential. Section 5 concludes with a summary of the findings and a discussion on the future directions this research could take, including potential advancements and the broader implications in the nuclear industry.

## 2. Background

This section provides an overview of the literature on integrating knowledge with data-driven techniques for PdM in the nuclear context. It begins (i) by **defining essential terms**, then (ii) explores **data-driven approaches for PdM**, outlining both their limitations and (iii) the **important role of knowledge-based approaches in enhancing these strategies**.

### 2.1. Basic definitions and concepts

A **fault** is an unauthorized deviation from the desired condition of a specific characteristic property of a system. This definition underlines that a fault represents a divergence from the norm that has not been permitted or planned within the system operations (Ma & Jiang, 2011). Detecting faults is crucial as they represent the initial indicators of potential operational disruptions. Within the spectrum of faults, distinctions are made based on their impact on system operations. A **blocking fault**, also known as a major fault alarm, is a severe defect that interrupts the continuation of operations and necessitates immediate corrective maintenance. This terminology aligns international standards (ISO 13372, 2012; ISO 14224, 2017) which define and classify fault types based on severity and operational impact.

Conversely, a **non-blocking fault**, often referred to as minor fault alarm, involves intermittent alarm that, although not halting the system function, requires acknowledgment by technicians each time they occur. While these faults are less critical, their persistent nature may be disruptive if not managed efficiently, and they may escalate if repeatedly ignored. Such faults are commonly classified in industrial maintenance standards (ISO 13372, 2012) as minor anomalies

Progressing from a fault, a **failure** is defined as the permanent cessation of a system ability to perform its required function within specified performance criteria (ISO 13372, 2012). Essentially, a failure occurs when the accumulative impact of unresolved fault reaches a critical threshold where the system can no longer sustain operational requirements. This irreversible state highlights the critical need for effective fault management strategies to preempt such failures, particularly in critical environments like nuclear facilities where the consequences can be dire.

**Data-driven techniques** use the advanced sensor infrastructure of Industry 4.0 to analyze patterns in data that predict these failures before they occur. These techniques are highly effective in environments rich in operational data, enabling models that can foresee and mitigate potential disruptions (ISO13379, 2015).

Simultaneously, **knowledge-based techniques** incorporate established expert knowledge, rules and principles derived from domain-specific expertise into the decision-making process. These techniques often use knowledge bases, including facts, heuristics, and logical rules, to simulate human decision-making or problem-solving processes (Chiang et al., 2001). Knowledge-based systems are designed to leverage the accumulated experience and insights of experts, providing guidance or solutions in areas where data may be sparse, incomplete, or too complex for purely data-driven approaches.

**Failure prediction**, a key objective in PdM, involves the use of both data-driven or knowledge-based approaches to estimate the future condition of system components and their remaining useful life. This proactive stance not only identifies the current state of system health but also predicts future degradation, facilitating timely interventions to extend the lifespan of critical components (ISO 13372, 2012).

### 2.2. Related work on data-driven methods for PdM and their limitations

Data-driven techniques designed to predict failures for enhanced maintenance planning have attracted substantial interest among researchers. This interest is evidenced by an extensive corpus of research, which includes 1686 articles indexed on Web of Science using the keywords "data driven" and "predictive maintenance" from 2015 to 2025. Such a considerable volume of research highlights the critical significance of this field (Si et al., 2011).

The classification of data-driven approaches in PdM is notably diverse, with no widely accepted consensus in the literature (Huang & Gao, 2020; Katerina et al., 2020; Mrad & Mraihi, 2023). Studies mainly include :

- **Statistical models:** These methodologies are among the earliest data-driven techniques used for predictive maintenance. They rely on historical data to identify trends, patterns and correlations to predict future equipment failures (Si et al., 2011). This primarily includes regression modeling on one hand. For instance, Sadeghi and Askarinejad (2010) propose a regression analysis in an exponential form to establish a correlation between some effective parameters and track degradation. Similarly, Chen et al. (2023) employ a linear regression model to capture equipment degradation and aging, as well as predict equipment health from operational parameters. Data in these studies often come from events such as maintenance actions or logs from equipment monitoring systems (Sadeghi & Askarinejad, 2010). Time series models, on the other hand, use physical measurements to train various forecasting models including Autoregressive Integrated

Moving Average (ARIMA), Autoregressive (AR), Moving Average (MA) or Box-Jenkins models (Facchinetti et al., 2022; Ganga & Ramachandran, 2020). Cancemi et al. (2025) extended this line of work by integrating ARIMA forecasting with deep learning-based health indicators for predictive maintenance in nuclear power plants.

- **Probabilistic Models:** incorporate uncertainty into their predictions employing statistical techniques to estimate the likelihood of future events. These models are particularly well-suited for situations where data is incomplete or inherently stochastic, and they often rely on probability distributions. Bayesian inference is a notable example of such techniques. For instance, Kapuria and Cole (2023) integrates survival analysis (Kaplan–Meier and Weibull) with Bayesian statistics to predict the health of a pump based on its current condition. Another probabilistic approach use Hidden Markov Models (HMMs) to forecast the future state of equipment (Vrignat et al., 2015). In the nuclear sector, Zanotelli et al. (2024) apply Left-Right Gaussian HMMs for condition-based prognosis, taking into account the effects of successive repairs on system reliability. Additionally, Tran et al. (2012) propose a three-stage method, combining Autoregressive Moving Average models with a prognostic framework based on Cox proportional hazard model to capture the system survival function. This model is then used to predict the Remaining Useful Life(RUL) of the component which refers to the estimated time duration that a machine or component is expected to operate before failure or requiring maintenance. RUL estimation is a critical aspect of predictive maintenance, enabling timely interventions and reducing unexpected downtimes. Dong et al. (2025) introduced a probabilistic RUL estimation method using Kernel Density Estimation (KDE). This approach enables dynamic modeling of RUL distributions and provides uncertainty-aware prediction intervals—particularly relevant for critical maintenance planning in industrial systems.

- **Machine Learning(ML) models:** ML models use algorithmic techniques to learn from data, automatically adjusting their parameters and improving predictions with new data. They handle complex, non-linear relationships in large data sets and adapt dynamically to new patterns. Key types include Deep Learning models, which utilizes advanced neural networks with deep architectures to capture intricate data patterns (Kizito et al., 2021; Pierleoni et al., 2022; Yang & Liao, 2024). For example, Luk et al. (2025) applied a CNN-LSTM encoder–decoder to forecast a constructed Health Index demonstrating its effectiveness in short-term maintenance prediction. Similarly, Al-Refaie et al. (2025) used an Multi-Layer Perceptron regressor for RUL prediction of cutting tools. K-Nearest Neighbors also offers a straightforward yet effective approach for classification and regression by averaging the features of the nearest data points (Deepika et al., 2025). Support Vector Machines are particularly effective in high-dimensional spaces and are robust against overfitting, especially beneficial when the number of dimensions exceeds the number of samples (Zenisek et al., 2019). Tree-based models, such as decision trees, Random Forest, or extreme gradient boosting algorithms, enhance predictive accuracy and control overfitting (Aizpurua et al., 2019). For instance Alshboul et al. (2024) compare seven ML classifiers for failure prediction in concrete manufacturing, with CatBoost achieving an F1-score of 0.985, illustrating the potential of ensemble models in PdM.

Despite their predictive capabilities, these data-driven approaches present several limitations. Table 1 provides a comparative summary of data-driven approaches in failure prediction, highlighting their respective strengths and limitations.

The efficacy of PdM models largely depends on the prediction objective whether the goal is to anticipate failures within a specified time horizon or to estimate the RUL of monitored systems. Although data-driven approaches have demonstrated strong predictive capabilities, particularly in high-dimensional and nonlinear contexts, they also face several limitations in safety-critical and complex industrial settings, such as the nuclear industry. As emphasized by Shukla et al. (2022), these models often fail to incorporate the underlying physical behavior of systems or expert knowledge, which limits their robustness and applicability in practice. These limitations can be categorized into three key areas: dependence on data quality, lack of interpretability and generalization issues.

First, many of these models are *highly dependent on data distribution and quality*. As emphasized by Wang et al. (2025) and Nie et al. (2025), their performance can degrade significantly under noisy, uncertain, or evolving operating conditions. This sensitivity makes them fragile in real-world scenarios. Similarly, Li et al. (2024) highlight the challenges faced by data-hungry prognostic models in industrial settings, underlining the importance of data augmentation. In this context, the integration of domain knowledge is essential to guide and validate the augmentation process.

Second, interpretability remains a major challenge. While ML models can deliver accurate predictions, they often lack interpretability, which limits their usefulness for informed decision-making. As highlighted by both Bouhadra and Forest (2024) and Gawde et al. (2024), models that do not provide transparent reasoning are difficult to integrate into operational processes. Post-hoc explainability tools have been introduced to address this issue; however, since these operate externally to the model itself, they do not fully close the trust gap between the system predictions and expert acceptance.

Finally, *generalization across diverse systems remains problematic*. As discussed by Yan et al. (2025), even advanced architectures like graph-based neural network models suffer from overfitting when applied to imbalanced or limited datasets. Beyond data-related challenges, their lack of grounding in physical principles and expert knowledge limits their applicability in high-risk environments.

Reflecting this view, Peng et al. (2010) and subsequent works (Koksal et al., 2024; Mokhtarzadeh et al., 2024; Ubesigha et al., 2025) advocate for hybrid approaches that combine data-driven learning with domain knowledge to enhance robustness, transparency, and relevance. Integrating domain knowledge — not only during data preprocessing (e.g., cleaning, feature engineering, integration) to improve data quality and relevance, but also in guiding model implementation and interpretation — provides a promising path toward more robust, interpretable predictive models, particularly in complex and heterogeneous industrial systems.

In summary, while data-driven models are powerful, their effectiveness remains limited by challenges related to data quality, interpretability, contextualization and generalization. A structured hybridization with domain expertise can help overcome these limitations and significantly improve the reliability and usability of predictive maintenance systems in critical applications (Klein, 2025).

## 2.3. Related work on knowledge enhanced data-driven in PdM

Knowledge-based models can play a pivotal role in optimizing maintenance equipments. These models are constructed upon a generic concept that encompasses a holistic consideration of the outcomes of maintenance activities, the condition of the system, and the organizational and procedural context. According to Nemeth et al. (2018), the essence of knowledge-based models lies in their ability to offer a comprehensive framework for enhancing maintenance strategies through the integration of various factors that influence maintenance decisions. A knowledge-based system operates by utilizing a knowledge base, which serves as a repository for the symbols of a computational model, represented in the form of domain-specific statements. Abdelillah et al. (2023) elaborate on the functionality of such systems, noting that they engage in reasoning by manipulating these symbolic representations.

Knowledge-based approaches are categorized into knowledge graphs and ontologies, rule-based systems, fuzzy systems and case-based reasoning (Nunes et al., 2023) :

**Table 1**
Comparative summary of data driven techniques.

| Data-driven techniques in failure prediction | Strengths | Weaknesses |
|---|---|---|
| Statistical approaches Cancemi et al. (2025), Diversi et al. (2025), Ekpenyong and Udoh (2024) | • Interpretable and transparent (e.g., regression, ARIMA,..) <br> • Effective with small dataset <br> • Easy to implement and computationally efficient | • Limited in capturing complex nonlinear relationships <br> • Requires domain knowledge for model selection and assumptions <br> • Often assumes stationarity or linearity <br> • Less robust to noisy or high-dimensional data |
| Probabilistic approaches Ferrisi et al. (2025), Wei et al. (2025), Zanotelli et al. (2024), Zhang et al. (2024) | • Captures uncertainty in predictions (e.g., Bayesian models, HMM, survival analysis,etc.) <br> • Provides confidence intervals and RUL distributions <br> • Flexible with incomplete or censored data | • Can be computationally intensive (e.g., variational inference) <br> • Requires good prior knowledge <br> • Model complexity grows with system complexity |
| ML approaches Elkateb et al. (2024), Li et al. (2025), Luk et al. (2025), Zhuang et al. (2024) | • Handles high-dimensional, nonlinear, and multi-source data <br> • Learns complex degradation patterns automatically <br> • High predictive accuracy | • Often lacks interpretability <br> • Requires extensive domain knowledge to contextualize results and to compensate for small or sparse datasets. <br> • Requires large amounts of labeled data <br> • Can overfit or underperform under data imbalance |

- **"Knowledge graphs"** often intersect with **"ontologies"**. An ontology is rigorously defined as "a detailed representation or specification of a conceptualization for a domain of interest" (Ayadi, 2018). It requires the use of formal logic to clearly model concepts and their interrelations, ensuring a structured and unambiguous representation of domain knowledge.
- **In rule-based systems,** knowledge is encapsulated within "if-then" rules hand-coded based on interviews of domain experts. These systems are structured around three core components: a knowledge base that contains the rules, a facts base that archives inputs, and an inference engine that applies the stored rules to the inputs in the facts base to generate new facts (Ayadi, 2018).
- **The fuzzy-knowledge-based models** extend the principles of rule-based systems through the incorporation of fuzzy logic. Unlike traditional 'if-then' rules that operate on binary true or false values, fuzzy logic allows for reasoning with partial truths. This means truth values can exist on a spectrum between absolute truth and falsehood (Abdelillah et al., 2023; Nunes et al., 2023). For example, the condition of equipment can be quantified as being 100% in good condition, gradually decreasing to 50% as wear begins, and ultimately reaching 0% when a fault occurs. This nuanced handling of truth values closely mirrors human reasoning, making fuzzy logic particularly adept at managing the complexity and ambiguity inherent in real-world scenarios (Gharib & Kovács, 2024).
- **The Case-based Reasoning approach** leverages previous experiences or cases to solve new problems. The decision-making process within these systems is driven by the evaluation of similarities between new observations and a database of previously documented situations. This approach enables the system to determine the most appropriate actions based on past experiences and accumulated knowledge (Klein, 2025). By comparing current situations to historical instances, case-based reasoning models can identify patterns and suggest solutions based on past outcomes. This model emphasizes the importance of experiential knowledge, highlighting a direct application of existing knowledge to PdM tasks (Nunes et al., 2023).

Recent scholarly work underscore the efficacy of hybridizing knowledge-based approaches with data-driven models to enhance PdM

strategies. Several earlier contributions have demonstrated the benefits of this integration. For instance, Barry and Hafsi (2023) propose a hybrid method to predict the RUL of aircraft engines by integrating ML and deep learning models with knowledge-based insights, thereby significantly improving prediction accuracy. This method not only forecasts RUL with higher precision but also adapts dynamically to changes in engine behavior, providing a robust solution for aerospace maintenance needs. In the same line, Gay (2023) focuses on RUL estimation in data-scarce environments by incorporating domain knowledge into the data augmentation process. This approach ensures the generation of realistic, representative samples, which not only improves the training of prognostic models but also supports more informed maintenance decision-making within the PHM framework. Cao et al. (2020) advance this concept by refining rule-based systems, incorporating expert-derived rules with data mining techniques, and employing quality measures like accuracy and redundancy to boost decision-making efficacy. This refinement leads to a more efficient handling of complex datasets, resulting in faster and more accurate maintenance predictions in manufacturing settings. Additionally, Aboshosha et al. (2023) explore a PdM framework using Fuzzy Logic Systems and Deep Learning in IoT context, further pushing the boundaries of industrial maintenance strategies. This framework effectively processes and interprets deep learning outcome, facilitating timely interventions that prevent costly equipment failures. More recent work continues to push the boundaries of knowledge-enhanced learning for PdM. Rajaoarisoa et al. (2025) propose a hybrid framework for wind turbine maintenance that integrates machine learning with a rule-based explanation layer, delivering accurate RUL prediction while providing interpretable insights to guide human decision-making. Similarly, Wang et al. (2025) present a supervised soft sensing approach that embeds domain knowledge into a hybrid graph neural network. Applied to a debutanizer column, the method enhances prediction accuracy and interpretability by capturing spatial–temporal process dependencies—highlighting the value of model-centric knowledge integration.

Collectively, these studies demonstrate the potential of combining knowledge-based techniques with data-driven models for predictive maintenance across various sectors, including aircraft,electronics, aerospace, agriculture, and manufacturing. This convergence aligns with the broader of PHM techniques, which emphasizes the importance of accurate, interpretable, and actionable predictions in industrial systems. However, a closer analysis reveals several important gaps in the way this integration is currently implemented.

First, *knowledge is often incorporated in a limited and ad hoc manner*, typically restricted to specific stages of the modeling process (Cai et al., 2024; Wang et al., 2025). In some cases, it is applied at the data level, to improve feature engineering or data quality; in others, it is used at the modeling level, to guide algorithm choice or improve interpretability. Frequently, it is only introduced after prediction, through explainable AI tools intended to clarify black-box outputs. While these strategies have value, they remain external to the model internal logic and contribute little to the robustness of the learning process itself.

Second, there is *a lack of structured, reusable methodologies* capable of integrating both explicit and tacit knowledge in a consistent and formalized way. Many existing works rely on isolated elements — rules, graphs, physical models — used on a case-by-case basis, often without a standardized or generalizable framework (Cao et al., 2020; Cummins et al., 2024; Kasilingam et al., 2024). As a result, their transferability across applications and domains remains limited.

Third, and most critically, *the nuclear sector remains largely underexplored in this regard*, despite its strong dependence on expert knowledge, strict safety standards, and abundant historical documentation (e.g., failures modes, effect and criticality analysis (FMECA) document, maintenance logs, operational procedures). The potential of systematically embedding this domain expertise into predictive ML models to support efficient PHM has yet to be fully realized.

To address these gaps, this work proposes a global and structured hybrid methodology that integrates domain knowledge throughout the entire PdM workflow—from data preprocessing to failure prediction modeling. Based on formal knowledge engineering principles, the approach captures both explicit and tacit knowledge and embeds it into the modeling process. Applied to nuclear equipment, it enhances data quality, improves model selection, and strengthens interpretability—laying the groundwork for more reliable, robust, and trustworthy predictive maintenance systems.

## 3. Proposed knowledge- and data-driven methodology

This paper proposes a hybrid methodology to support PdM in the nuclear industry, with a specific focus on failure prediction—a core function of PHM. The approach combines domain knowledge (explicit and tacit) with data-driven ML models. It is structured in four main stages, integrating knowledge throughout the workflow: from data preprocessing to model training, evaluation, and deployment, as illustrated in Fig. 1. Domain expertise improves data quality, data augmentation, supports failure prioritization, and enhances both model guidance and interpretability. This integration aims to leverage the complementary strengths of knowledge engineering and ML for robust, interpretable, and context-aware failure prediction. Each of these stages is detailed in the following subsections.

### 3.1. Stage 1: Knowledge engineering

Structured knowledge is extracted from domain experts and historical documentation (e.g., FMECA, maintenance reports). This forms a reusable knowledge base that supports the downstream tasks.

The nuclear industry is endowed with rich knowledge, encompassing deep expertise in industrial equipments that are crucial for online monitoring and maintenance. Stage 1 aims to capture and analyze this knowledge. Knowledge engineering, a field at the intersection of cognitive science and socio-cognitive engineering, involves the generation and organization of knowledge by socio-cognitive entities (primarily humans), structured around the understanding of human reasoning and logic (Suebsombut, 2021). Nuclear equipment expertise offers a detailed understanding of system operations, including its failures (alarms, faults, failure). Maintenance expertise, on the other hand, provides knowledge into breakdowns (corrective actions) which, depending on the maintenance intervention priority, may significantly affect production. This expertise extends to preventive maintenance actions, like periodic controls and diagnostics, documenting various interventions on the equipment. The knowledge may be explicit (i.e., documented) or tacit (i.e., acquired from discussions with experts). The Knowledge Engineering phase begins by preprocessing this diverse knowledge — including PDFs or Excel files, oral recordings or graphical data — converting them into a format suitable for further analysis. The process is iterative. At the end, a model is developed to accurately represent various operational scenarios: alarm situations, minor faults, blocking faults requiring maintenance actions, and states of normal operation. The model outputs are converted into knowledge base to facilitate integration into subsequent stages.

### 3.2. Stage 2: Data preprocessing (knowledge informed data)

The extracted knowledge is used to improve data quality, identify outliers, and enrich features from raw sensor and operational data. This ensures that the processed data better reflect the actual monitoring conditions of the equipment.

The success of ML models largely depends on the quality and form of the input data. Addressing challenges such as handling missing values, filtering outliers, and correcting time-related anomalies are crucial, problem-specific data preparation tasks (Zenisek et al., 2019). The four phases for preprocessing are outlined in Fig. 2.

**Phase 1: Data cleaning**. The input data combine physical measurements (e.g., temperature, pressure) recorded as time series, with annotated fault histories extracted from supervisory systems. Data cleaning is adaptive and case-specific, driven by knowledge extracted from the stage 1. The contribution of domain knowledge to the main components of data cleaning is detailed below.

- Missing value treatment: The labeled knowledge created in stage 1 plays a central role in this step. For example, When missing values are due to automation disturbances or system malfunctions, they are reconstructed using past failure logs and knowledge-defined behavior patterns. On the other hand, values missing due to scheduled shutdowns are excluded, since they are not informative for failure prediction. Based on the operational context, the knowledge base helps determine the most appropriate strategy whether deletion, interpolation, or estimation ensuring that the process is consistent with real system behavior. This ensures that the treatment of missing values enhances the accuracy and reliability of the overall analysis, while aligning with the broader goals of the data processing methods.
- Outliers detection: Outlier detection is informed by knowledge that define what constitutes an "outlier" under various operating scenarios. This step distinguishes between erroneous data (false positives) and valid, informative anomalies (true outliers). It involves:

  - *Identifying and correcting false positives*: Sensor readings can sometimes indicate faults where none exist, due to environmental interference or momentary disturbances. The knowledge base enables the identification of such cases based on known patterns and fault labeling, allowing to systematically filter out non-relevant data and prevent misinterpretation.

  - *Dealing with true outliers*: Outliers that remain are evaluated based on their relevance to the prediction objective. The knowledge base helps determine whether the deviation corresponds to a meaningful degradation pattern or an isolated event. Non-relevant points are excluded, while those linked to failure precursors are retained and flagged to enhance the learning process.

## METHODOLOGY

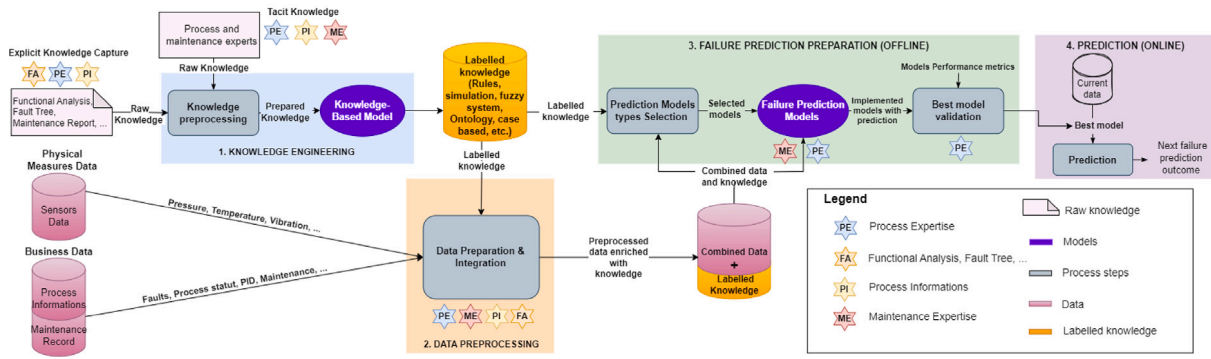

**Fig. 1.** Proposed hybrid methodology for failure prediction in nuclear equipments.

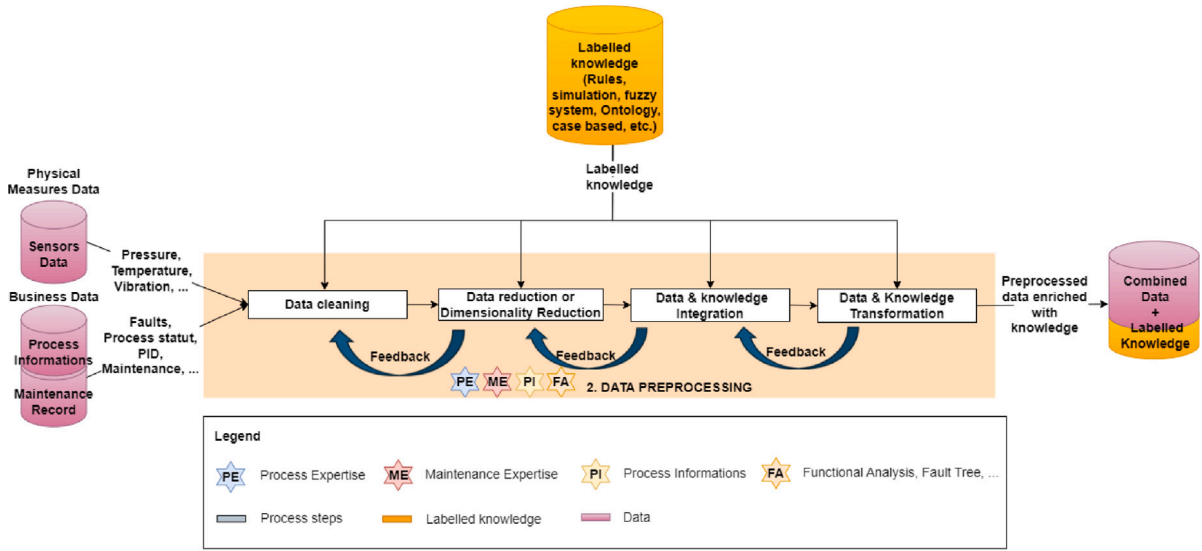

**Fig. 2.** Data preprocessing in Stage 2.

At each step, the structured knowledge base is not simply supportive—it is central. It ensures that cleaning decisions are grounded in operational logic, domain understanding, and historical behavior, rather than relying on generic statistical thresholds.

***Phase 2: Data reduction or dimensionality reduction.*** Following the cleaning phase, the data reduction stage focuses on isolating relevant data for prediction. Techniques used include feature selection and the row-wise data sample reduction. The contribution of domain knowledge to each of these processes is detailed below.

- Feature selection: Selecting the most informative variables allows predictive models to concentrate on factors that directly influence failure behavior. While standard ML or data mining techniques are applied to identify high-impact features, the knowledge base plays a central role in validating their relevance. This validation ensures that selected features align with the failure modes and operational realities of the equipment, making the input space both meaningful and domain-consistent.
- Row-wise data sample reduction: In cases involving large datasets, reducing the number of samples may improve computational efficiency. Here again, the knowledge base guides the selection process, helping ensure that reduced datasets remain representative of critical operational scenarios and failure signatures. This step safeguards the model's exposure to relevant patterns while reducing noise and redundancy.

***Phase 3: Data and knowledge integration.*** Once relevant data are isolated, this phase focuses on integrating structured knowledge as features into the dataset. The result is a curated and enriched dataset that combines refined time-series measurements, faults logs with knowledge-defined indicators and fault-specific information. The knowledge base contributes to this integration in three key ways:

- Normal operating parameters: are directly extracted from the knowledge base. These parameters reflect expected equipment behavior under nominal conditions and are used to define meaningful thresholds for anomaly detection. These thresholds allow normal behavior to be explicitly tagged in the data, enabling to distinguish between nominal and abnormal conditions. By relying on predefined expert knowledge, this integration ensures that deviations in the data are interpreted in line with real-world operational expectations.
- Failure characteristics: Detailed information about failure modes, their causes, effects, and criticality — extracted from the knowledge base (e.g., FMECA) — is added to the dataset. This enables a better understanding of how specific faults manifest and evolve, thereby improving the quality of predictive labeling and model learning.
- Predictive failures refinement: Faults are prioritized based on their frequency and operational impact. Using knowledge-defined criteria (e.g., Pareto analysis), the integration process emphasizes the most critical failure types. This ensures that predictive models are trained with a focus on high-risk events, improving both relevance and resource allocation in downstream maintenance decisions.

***Phase 4: Data and knowledge transformation.*** This phase transforms the enriched dataset into a form suitable for use with machine learning models. The transformation process adapts flexibly to the characteristics of the case study and the selected algorithms. Domain knowledge continues to play a guiding role throughout this phase, particularly in the following aspects:

- Numerical data transformation: Numerical variables are processed using methods such as normalization, standardization, or binning (equal-width or equal-frequency), depending on model requirements. The choice of transformation is informed by domain-specific knowledge to ensure consistency with the operational interpretation of each variable (e.g., critical thresholds, ranges of safe operation).
- Categorical data transformation: Categorical inputs — whether sensor states, operating modes, or knowledge-based labels — are encoded using appropriate techniques. knowledge feature are also converted into categorical inputs and structurally integrated into the dataset. This ensures that symbolic domain knowledge is preserved and made usable by the model.
- Imbalanced data handling: When certain failure types are underrepresented, the dataset is rebalanced using knowledge-guided techniques. Depending on the fault criticality, historical occurrence, and system behavior, the knowledge base helps define suitable resampling or augmentation strategies. These include targeted sampling near specific failure events, synthetic data generation, or the application of weighted class schemes. This knowledge-informed approach ensures that minority fault classes are properly represented while preserving the integrity of real operational data, improving the model's robustness and generalization capacity.

### 3.3. Stages 3 & 4: Failure prediction

The structured knowledge guide the implementation of suitable ML models. The best-performing model is then identified based on relevant performance metrics in stage 3. Stage 4 deals with online deployment and continuous learning.

Stage 3 of the methodology in Fig. 1 focuses on the offline failures prediction over a specified horizon. This stage leverages the curated dataset and structured labeled knowledge obtained in the previous phases. The following outlines how domain knowledge is incorporated throughout this stage. Fig. 3 provides a global view of the process.

The stage begins with the selection of predictive models, guided by the prediction objectives, the structure of the curated dataset, and the nature of the failures to be predicted. Domain knowledge plays a critical role in this phase. It helps define the prediction horizon, select relevant input variables, and determine whether the modeling task involves classification (for discrete fault events) or regression (for continuous degradation trends). Once appropriate models are selected, they are implemented and evaluated. The dataset is partitioned into training, validation, and testing sets according to standard ML practices. However, this step is enriched by domain-informed partitioning— ensuring that the split reflects realistic operational scenarios and the temporal progression of faults:

- The training set is used to expose the models to diverse failure behaviors, enabling them to learn relevant patterns through repeated examples.
- The validation set supports hyperparameter tuning, where performance is monitored to optimize the configuration of each model.

- The test set provides an unbiased evaluation of the final model, measuring generalization performance on unseen data.

Throughout this process, the knowledge base continues to inform key decisions. It guides the definition of relevant features, highlights critical fault types that should not be overlooked during training, and supports the interpretation of model performance. Additionally, the selection of evaluation metrics is tailored based on the failure criticality and operational constraints. Once training and evaluation are complete, the model that offers the best balance between predictive performance and generalization capability is selected. This best candidate demonstrates not only high performance metrics on the training and validation sets but also performs well on the test set over the targeted prediction horizon.

In Stage 4, the selected model is deployed in a real-time environment to anticipate failures ahead of time. Its outputs are translated into operational indicators and made available to experts for informed decision-making. In parallel, a feedback loop is integrated into the process. Expert input from equipment operators and maintenance teams is collected during deployment to assess the relevance of predictions and refine the knowledge base. This feedback may include confirmations or adjustments of predicted faults, additional contextual observations, or new failure patterns. These insights are used to update the modeling. This continuous learning mechanism allows the predictive system to evolve over time, improving its robustness, accuracy, and consistency with real-world operational conditions. By coupling predictive performance with knowledge-informed interpretability, the system delivers actionable insights that strengthen proactive maintenance planning in nuclear operations.

## 4. Case study

To validate the proposed methodology and assess its performance, a real-world case from the nuclear industry is examined. This section begins by introducing the equipment, detailing its operational mode and significance. Next, the current state of the equipment is evaluated with respect to PdM and compare it with two distinct fault prediction scenarios:

- Scenario 1 is a purely data-driven approach that leverages physical measurement data, historical equipment faults records and sequences information to train ML models for fault prediction.
- Scenario 2 applies the proposed methodology.

Following this, the section presents and discusses the outcomes of these scenarios, aiming to determine which scenario provides higher predictive performance. This case study is designed to demonstrate the potential of the methodology to enhance fault prediction within the nuclear industry, highlighting: (i) The methodology ability to analyze the overall operational experience of the equipment, (ii) Its adeptness at preprocessing diverse data types while incorporating domain knowledge, (iii) its capacity for predicting faults using data-driven techniques augmented by knowledge. Please note that due to **confidentiality** constraints, the actual names of these variables are anonymized in this article.

### 4.1. Case study description

the case study focuses on a critical equipment at a nuclear facility. This complex equipment, monitored by an automation control system, primarily functions to sample nuclear material.

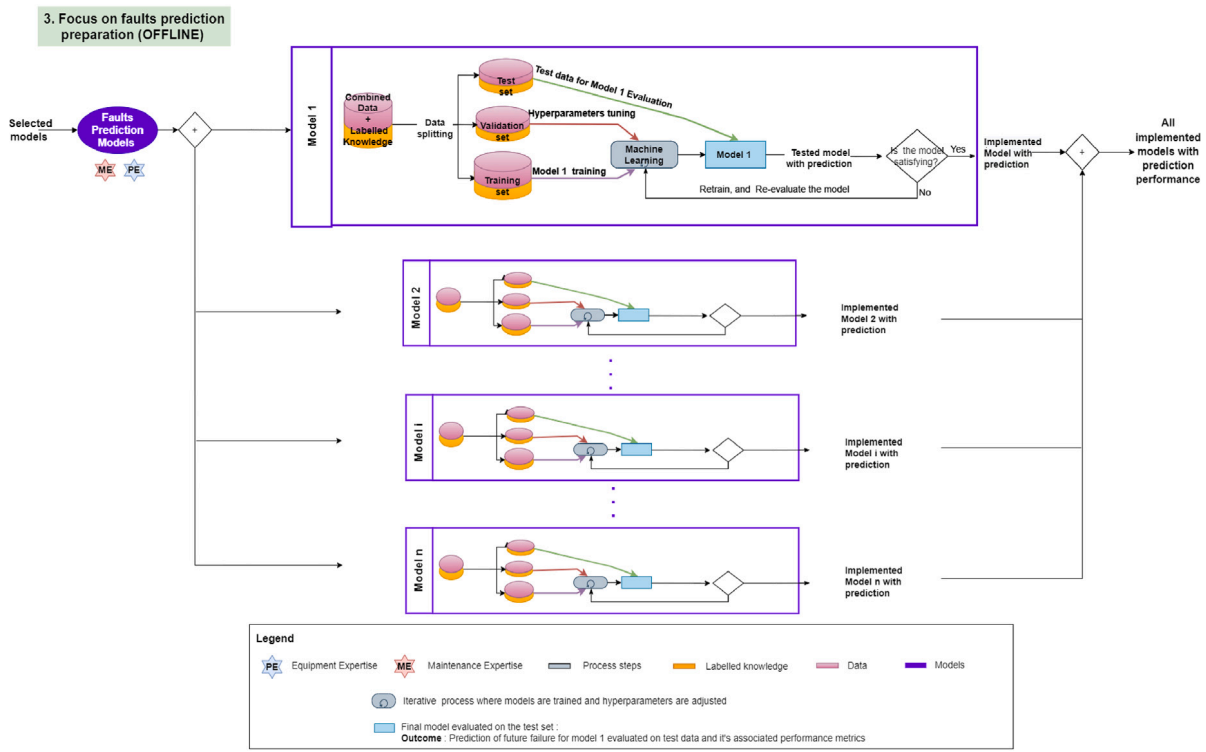

**Fig. 3.** Failure prediction in Stage 3: Models selection, ML models implementation and best model selection based on performance metrics.

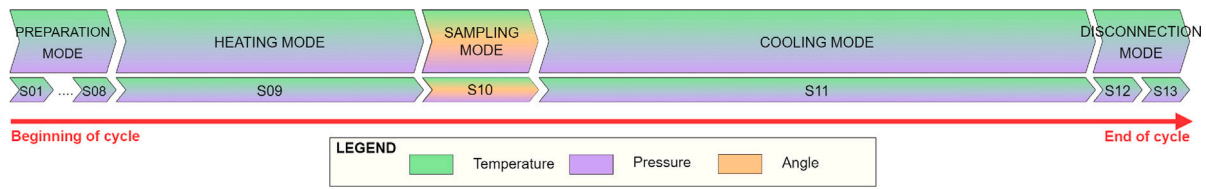

**Fig. 4.** Use case modes understanding.

### 4.1.1. Operation of the considered equipment

The equipment operates in a cyclic manner and is characterized by five principal operational modes, each comprising a set of sequences. Its full operational cycle is shown in Fig. 4 which illustrates the modes and sequence from the beginning to the end of a cycle associated with nominal duration:

- **Preparation mode**: Initiated manually by operators, this mode consists of eight sequences (S01 through S08). This phase sets up the equipment, including tasks such as connecting it to the control system and conducting connectivity and seal tests.
- **Heating mode**: Automatically triggered by the control automation system, this mode heats the equipment material to prepare the sampling. It includes a single sequence (S09).
- **Sampling mode**: Also automated, is initiated for the collection of samples and includes only one sequence S10.
- **Cooling mode**: This final active phase involves cooling the processed material. It is the longest phase of the cycle and includes only sequence S11.
- **Disconnection mode**: This mode marks the end of the operational cycle and involves the disconnection of the equipment. It comprises two sequences S12 and S13.

The purpose of this equipment is to sample in S10. A cycle is considered successful if it manages to carry out this sampling effectively within the nominal process time without any faults. Conversely, a sampling fault indicates a malfunction within the cycle, directly impacting the production capacity of the facility by increasing its downtime. The equipment is susceptible to two categories of faults: blocking and non-blocking. The equipment is continuously monitored by pressure, angle and temperature sensors (both internal and external). The data collected from these sensors check that the equipment operates within its designed parameters.

### 4.1.2. Objectif of the case study

the scientific goal is to ensure that the prediction is more accurate when ML is augmented not only with data but also with domain-specific knowledge. For this purpose, the objective is to predict the most critical blocking faults that could interrupt S10 before it begins, over a specific predictive horizon. For this study, the data and information collection covers the period of 2021 and 2022, period with a higher incidence of S10 faults. These data aggregate the following three main types of sources:

- **Sensors data**: This dataset primarily consists of pressure and temperature readings, complemented by angle measurements of the equipment. A dataset comprising 865,715 samples across 21 variables, with a 1 min sampling interval, is collected. The physical measurements capture both internal and external conditions of the equipment, providing a holistic view of its operational environment.
- **Faults logs**: The fault dataset includes records of all S10 faults logs that have occurred, serving as the prediction target. When

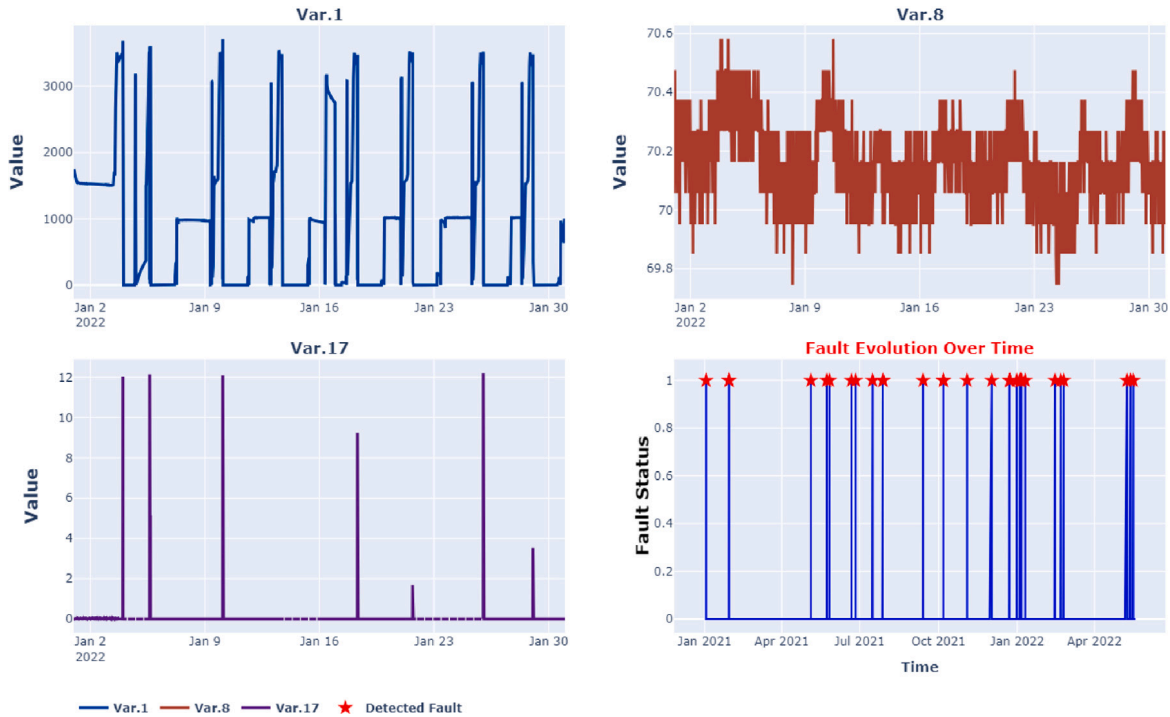

**Fig. 5.** Fault log and data of three sensors.

a fault occurs, the automation system displays "1" and automatically switches the equipment to offline mode until a technician intervenes. This intervention may involve corrective maintenance or simply restarting the equipment manually. After this, the system is reset to "0" indicating that the fault has been resolved and the equipment is ready for operation again.

- **Sequences logs:** This data captures the sequence of operations that the equipment follows from the start to the end of each cycle. Two variables are recorded: sequence logs and cycle number. These variables are integrated with physical measurements to monitor and control the expected values at each stage of the process. For the analysis, a total of 55 cycles are collected.

Fig. 5 provides an overview of 3 out of the 20 sensors variables for January 2022 and an overview of fault data. This overview of the sensor and fault data allows to understand the monitoring parameters of the equipment and their evolution. The following subsection will explore how the equipment is currently monitored and what this process entails, before diving deeper into Scenarios 1 and 2.

### 4.2. Current state of monitoring on the case study

The equipment is currently condition-based monitored, by employing threshold-based alarms configured by domain experts. Notably, no ML models are presently integrated into this monitoring process. That is, the thresholds are established from operational knowledge, documented through the functional analysis document. Fig. 6 illustrates how this operational knowledge (combined with continuous input from equipment and maintenance experts) is incorporated into the monitoring process.

Knowledge extraction from the functional analysis is performed by equipment and maintenance experts, who are directly responsible for designing a rule-based model which drives the equipment monitoring process. The extracted knowledge (K) and parameters (P) serve as the basis for configuring these rules.

The quantitative data used in the rule-based model are as follows:

- *Equipment mode*: Identifies the current operational state of the equipment
- *Sequence logs*: Specifies the particular step within the equipment mode
- *Sequence timing*: Indicates the duration or specific time associated with each sequence.
- *Sensors values*: Provides readings from various sensors according to the current mode and sequence.
- *Faults logs*: Record all faults occurring during the cycle, recorded in a binary format (1 for the appearance of a fault and 0 otherwise).

The rule-based model employs "if-then" conditions coming from the equipment parameters outlined earlier. These rules serve as the foundation for real-time decision-making, enabling a proactive approach to process management and fault prevention. To concretely illustrate how these rules are implemented by experts to monitor the equipment, Table 2 provides examples of a subset of the monitoring rules derived from the functional analysis.

### 4.3. Scenario 1: Pure data-driven approach

As previously mentioned, scenario 1 employs the methodology by omitting the knowledge engineering phase, thus relying on a purely data-driven approach. Before beginning the preprocessing, all raw data and logs informations including faults and sequence events are consolidated into one large database. This consolidation simplifies the subsequent stages and ensures a streamlined data handling process.

The remainder of this subsection details the application of the methodology presented in Fig. 1, without the integration of knowledge, to the case study.

#### 4.3.1. Stage 2: Data preprocessing

Unlike the data preprocessing described in the complete methodology, this scenario adopts standard preprocessing practices which include:

# CURRENT EQUIPMENT MONITORING: RULE-BASED APPROACH

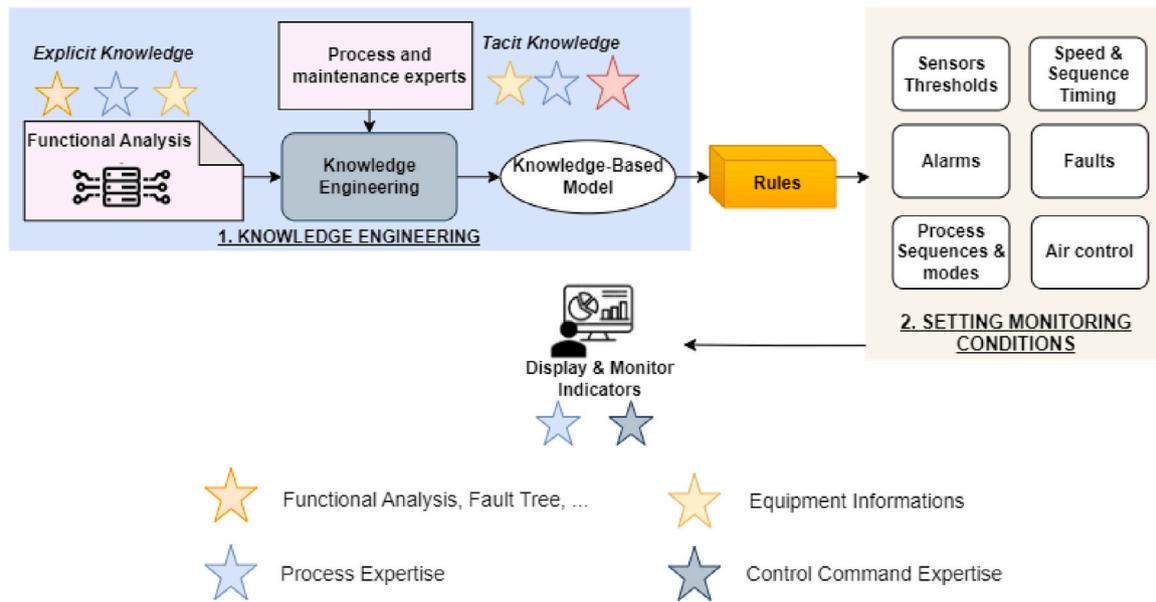

**Fig. 6.** Overview of the current equipment monitoring process.

**Table 2**
Examples of rules used in equipment monitoring.

| ID fault | Blocking/ Non-blocking fault | Modes | Sequences number | Sensors parameters | Logs | IF | THEN | Consequences |
|---|---|---|---|---|---|---|---|---|
| 1 | Blocking fault | Sampling | S10 | Pressure | Needle valve | **IF** there is no memory of "internal pressure < 50 hPa" during the first 5 min of S10 | Needle Valve Fault | Cycle stop: The expected pressure is not reached to proceed to the next steps |
| 2 | Blocking fault | Sampling | S10 | – | Valve 0001, Valve 0002 | **IF** (log valve 0001 OR valve 0002) = 1, at step 1 of S10 in the graph set | Sample Taking Fault | Cycle stop: Non-opening/closing of the valves during the sampling mode |
| 3 | Blocking fault | Heating | S09 | Temperature | – | **IF** temperature > 314 °C, at the step 5 of S 09 in the graph set | Heating fault | Cycle stop: The temperature exceeding the limit |
| 4 | Blocking fault | Heating | S09 | Pressure | Brewing fan | **IF** Pressure > 3141 hPa, at step 06 of S09 in the graph set AND log brewing log = 0 (Not running) | Heating fault | Cycle stop: The brewing fan not operating and pressure exceeds the limit |
| 5 | Blocking fault | Sampling | S10 | Angle | – | **IF** value ∉ [−31°, 40°] | Angle Measurement Fault | cycle stop: equipment inclination out of range |
| 6 | Non-Blocking fault | Preparation | S04 | – | Equipment door | **IF** log Z013 = 1 | Door Closure Fault | Alarm acknowledgment and resumption of the procedure: Closing the process door |

*Phase 1: Data cleaning.* This phase involves removing all missing values, non-use periods, and outliers from the dataset consisting of 865,715 samples across 24 variables. As shown in Fig. 7 for Scenario 1 data cleaning phase, missing values are identified and removed, resulting in the deletion of all observations containing Not-a-Number values, including variables where the missing values percentage exceeds 50%. Finally, statistical methods, such as invariant component selection outliers detection and box plot analysis, are used to identify and eliminate non-use periods and outliers. By the end of this phase, the cleaned database contains 790,000 observations across 23 variables.

*Phase 2: Data reduction.* In this phase, data reduction is achieved through data mining techniques, such as correlation analysis and Principal Component Analysis. These methods identify the most influential variables within the physical measurements along the principal component axes and correlation values. The selected variables are those that most significantly explain faults in the sampling sequence.

*Phase 3: Data integration.* Then, this phase combines the selected variables from the data and introduces additional statistical features (mean, median, and variance) that are relevant for time series analysis.

**Table 3**
Performance metrics.

| Criterion | Mathematical formula |
|-----------|---------------------|
| Accuracy | $\frac{TP+TN}{TP+FP+FN+TN}$ |
| F1 Score | $2 \times \frac{Precision \ \times \ Recall}{Precision+Recall}$ |

*Phase 4: Data transformation.* Finally, this phase of Scenario 1 prepares the data for ML modeling. It begins by standardizing all quantitative data to a uniform scale, ensuring that scale discrepancies do not bias the analysis. It then converts categorical variables into dummy variables using encoding techniques. This conversion makes these variables easily interpretable by ML models, enhancing the accuracy of the analysis. To address the issue of data imbalance which may skew predictive outcomes, it selects relevant observations that represent both typical and atypical operating conditions associated with S10 faults. The outcome of this transformation phase is an optimized and enriched dataset ready for faults prediction(Stage 3).

### 4.3.2. Stage 3: Sampling faults prediction

Stage 3 predicts sampling faults using three selected ML models: Random Forest, Light Gradient-Boosting Machine (LightGBM), and Support Vector Machine (SVM). These models were chosen based on their frequent application in PdM research and their proven effectiveness in handling industrial time series data (Allal et al., 2024; Carvalho et al., 2019; Chapelin et al., 2025; Xu et al., 2025). Random Forest and LightGBM are ensemble learning methods known for their robustness to noise, ability to model non-linear relationships, and strong performance with imbalanced datasets—common characteristics in industrial monitoring scenarios. SVM was included for its strong generalization capabilities in high-dimensional feature spaces and its popularity in PdM tasks requiring precise classification boundaries. Together, they cover diverse learning paradigms — bagging, boosting, and kernel-based methods — offering a representative and complementary basis for assessing the impact of domain knowledge integration. In addition, these models are widely used in safety-critical environments, including the nuclear sector (Liu et al., 2024; Saley et al., 2022).

Stage 3 sets the prediction horizon and partition the data into training (60%), validation (20%), and testing (20%) sets. Model performance are evaluated using accuracy and F1 score metrics, as well as the predictive horizon.

A minor data recalibration is performed to accommodate different temporal horizons, with window sizes set at 3 h, 12 h and 24 h. The evaluation of these models is consistent, applying performance criteria to select the most effective model. These criteria include achieving an accuracy greater than 70%, the highest F1 score among the three models, and the longest predictive horizon associated with these metrics. Table 3 provides the formulas of the performance metrics:

Where:

- $TP$ is the number of true positives.
- $TN$ is the number of true negatives.
- $FP$ is the number of false positives.
- $FN$ is the number of false negatives.
- Precision = $\frac{TP}{TP+FP}$ is the ratio of correctly predicted positive observations to the total predicted positives.
- Recall = $\frac{TP}{TP+FN}$ is the ratio of correctly predicted positive observations to all observations in the actual class.

### 4.4. Scenario 2: proposed methodology

This subsection applies the proposed methodology to the case study as follows:

### 4.4.1. Stage 1: Knowledge engineering

In this study, knowledge engineering involves extracting relevant information from both the functional analysis document of the equipment and the expertise provided by experts. Table 2 shows how the functional analysis maps the current monitoring rules of the equipment. This approach captures not only the standard operational parameters but also the specific conditions under which these parameters must be adjusted in response to emergency situations or anticipated failures. For example, rules include:

- *All equipment parameters*: all relevant parameters that can be monitored by sensors are detailed, establishing rules that define the normal operational ranges for each measured data to ensure consistent monitoring.
- *Fault types and causes*: a detailed classification of each type of equipment fault is detailed, including identifying potential causes such as mechanical wear, electrical failure, or operational errors.
- *Faults consequences and severity*: For each type of fault, its consequences are associated based on whether it stops the cycle (considered severe) or not, allowing to classify them into blocking and non-blocking faults.
- *Monitoring rules set up*: Rules are established to dictate the expected values or behaviors under normal operating conditions, including timing sequences and nominal profiles where applicable.
- *Corrective actions*: the actions to be taken when a fault is detected are specified, which may include immediate shutdowns, alerts to technicians, or other corrective measures to minimize risks (e.g equipment damage, production downtime, safety hazards, etc.).
- *Data during non-use*: Rules are set to monitor how the equipment should be managed when not active, which may include lower power modes and considerations for environmental conditions that could impact the equipment while idle.
- *Maintenance schedules*: Preventive and corrective maintenance schedules are integrated into the rules, featuring both time-based schedules and condition-based triggers that determine when maintenance should be performed.

Such above rules are stored in a knowledge base to facilitate the next stages.

### 4.4.2. Stage 2: Data preprocessing

This stage in Scenario 2 extends the four foundational phases present in Scenario 1 — data cleaning, reduction, integration, and transformation — by incorporating previously developed rules. Like in Scenario 1, it begins by consolidating all data and log information into a large database to simplify the preprocessing stage. This ensures that all relevant data is centralized, facilitating a more streamlined and efficient analysis process. The outcome of the data preprocessing is a comprehensive dataset that not only includes cleaned and integrated data but also features newly derived from the established rules.

*Phase 1: Data cleaning.* This phase addresses similar challenges as in scenario 1, such as handling missing values, accounting for non-operational periods, and detecting outliers. However, Scenario 2 integrates domain-specific knowledge. As in Scenario 1, raw data comprise of 865,715 observations across 24 input variables, which will undergo further preprocessing. Fig. 7 outlines the steps involved in data cleaning.

- Step 1 - Handling missing values & non use periods: this step detects missing values and consult the rules to determine their origins. There are generally two main cases as defined in the knowledge base:
  - When all sensors simultaneously fail to report data, coinciding with the major preventive maintenance periods (winter and summer), these observations are deleted from the dataset.

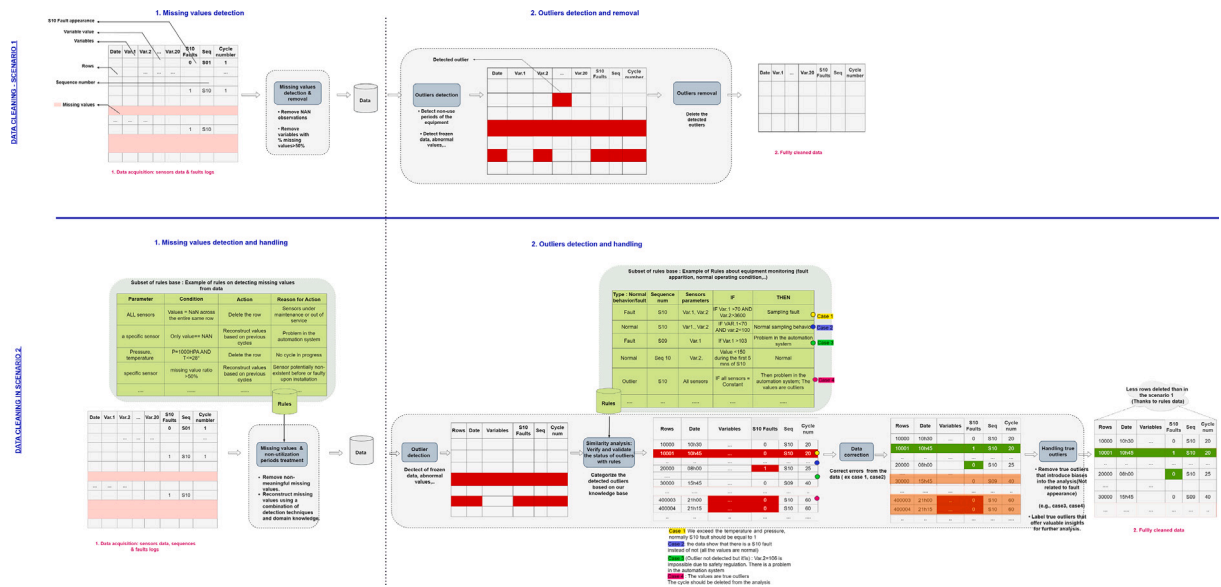

**Fig. 7.** Data cleaning steps in both scenarios 1 and 2.

– When a specific sensor reports missing values due to a loss of information from the data transmission system or if the sensor did not exist previously, these situations are handled by reconstructing the data by averaging observed data from the previous cycles.

Additionally, data from periods when the equipment is not in use are removed, indicating that there is no active operational cycle. This ensures that the dataset only includes periods of active usage.

· Steps 2, 3, 4 and 5 - Outlier detection, verification and correction: After handling missing values in phase 1 with a specific set of rules, it moves to steps 2 and 3 in which different rules are applied for detecting and correcting anomalies. These phases utilize techniques such as Invariant Component Selection Outlier, box plots, and the InterQuartile Range to identify outliers. Anomalies are detected as illustrated in red in the data cleaning (Fig. 7). The approach in these phases involves:

– Comparing detected anomalies against a different set of rules from the knowledge base, specifically tailored for anomaly correction. If discrepancies are found (as shown for example in cases 1 and 2 in Fig. 7), corrections are made.
– Tagging the data points where no discrepancies are found, confirming them as true anomalies, and considering them relevant for blocking fault prediction. True anomalies that are not relevant (cases 3 and 4 in Fig. 7) are removed, as they are confirmed to be associated with external factors unrelated to the equipment performance. Excluding these anomalies prevents potential bias in the fault prediction models, ensuring the analysis remains focused on relevant and actionable data.

By the end of this process, a database of 860,000 observations across 24 variables is compiled, more comprehensive than in Scenario 1 and prepared for the subsequent preprocessing steps. This enhanced approach ensures that the dataset not only maintains high quality and relevance but also aligns closely with the refined understanding and expectations from the data, setting a robust foundation for advanced analyses.

***Phase 2: Data reduction.*** After the data has been cleaned, this phase select only the key time series features important for the analysis to facilitate fault prediction.

Initially, the same analytical methods as Scenario 1 is used, using correlation matrices and Principal Component Analysis to select relevant features. Once selected, these features are compared against the knowledge base for validation. A variable is considered important if the associated sensor is not redundant and plays a role in every phase of the operational cycle, particularly in the sampling phase aiming to predict faults. In cases of variable redundancy, only one variable is considered for further analysis.

***Phase 3: Data & knowledge integration.*** This phase of preprocessing is important and involves integrating both knowledge-based and statistical features into the dataset retrieved from the previous phase. This addition enhances the explainability of the data, thereby facilitating efficient prediction of S10 faults. This phase includes many steps:

· Step 1: Addition of knowledge in data. First, rules derived from the knowledge engineering model are integrated as features to explain the faults observed in the data. When a fault is observed, three types of knowledge features are added to the data: (i) the cause of the fault appearance, (ii) the severity of the fault, and (iii) the consequences of the fault for the equipment. This step allows to represent the FMECA of the equipment in the data using the knowledge.
· Step 2: Addition of statistical features and faults prioritization feature. Once Step 1 is completed, statistical features are added to the variables as in scenario 1. Then, a fault prioritization feature is integrated: an exploratory analysis of S10 faults allows to isolate the most observed causes of S10 faults, thus enabling to refine the faults. The prioritization feature is binary and is set to 1 if the observed fault is severe (blocking) and is among the most severe faults, and 0 otherwise.
· Step 3: Reconstruction of prediction target (S10 faults) and its integration into data. Leveraging the in-depth equipment knowledge in identifying the root causes of critical faults, the prediction target is strategically reconstructed to precisely capture recurrent and high-impact fault scenarios. This approach moves beyond the limitations of relying on raw S10 fault data, as recorded by the automation system in Scenario 1. Instead, it focuses on a refined target that prioritizes faults with "blocking" severity, consequences leading to "cycle stop", and a fault prioritization level of 1, ensuring that the predictive model is both robust and closely aligned with the most significant operational challenges.

This structured approach ensures a comprehensive dataset that not only reflects the real-time operational status but is also enriched with domain-specific knowledge, thereby enhancing the predictive capabilities of the models.

***Phase 4: Data & knowledge transformation.*** This phase marks the last part of data preprocessing. It involves standardizing data to a uniform scale, converting categorical variables into a format suitable for analysis through encoding techniques, and generating derived variables to highlight operational patterns and anomalies.

The categorical variables identified in the previous steps are transformed into dummy variables, making them easily interpretable by ML models. As in Scenario 1, data imbalance is addressed by selectively including observations that most accurately represent the conditions of the target variables. Specifically, for each cycle, the data and knowledge from sequence S10 as well as the heating sequence immediately preceding S10 are included. This choice is based on data exploration and knowledge, which reveal that most faults occur during this time period. This targeted inclusion ensures that the dataset is both representative and robust.

### 4.4.3. Stage 3: Sampling fault prediction

This subsection delves into the implementation and evaluation of ML models to predict sampling faults from the knowledge-enhanced preprocessed data. To maintain consistency with the methodology and facilitate comparisons with Scenario 1, a focus on three critical phases is made:

- **Model Selection:** the reconstructed prediction target is also binary, necessitating the use of supervised classification models. Thus, the same three supervised classification ML models are used as in Scenario 1: Random Forest, LightGBM, and SVM. This approach ensures direct comparability of the outcomes.
- **Fault Prediction model implementation and validation:** The ML models are implemented following the same procedure outlined in the fault prediction step of Scenario 1. However, for Scenario 2, the models leverage the enriched dataset that integrates domain-specific knowledge.
- **Best Model Selection:** The optimal predictive model is selected based on accuracy and F1 score, which are critical metrics for assessing model performance. The preferred model is as in Scenario 1 is the one that not only forecasts an S10 fault with a significant lead time but also achieves an accuracy of at least 70% and the best F1 score compared to the other models.

### 4.5. Results

This section presents the outcomes of the two scenarios and compare their results against the current state of the equipment in terms of PdM. It begins with an overview of the data preprocessing results for both scenarios, followed by the results of fault prediction sampling. Finally, it concludes by comparing the predictive performance of both scenarios with the current state of the equipment monitoring system.

### 4.5.1. Data preprocessing results

This subsection provides a detailed overview of the results obtained from Scenario 1 and Scenario 2. Both Scenario 1 and Scenario 2 follow similar preprocessing methodologies, but Scenario 2 is distinguished by the integration of additional knowledge from the existing monitoring model. Scenario 1 focuses on traditional data preprocessing, including the detection and removal of missing values, identification and elimination of outliers, and reduction of data dimensionality using PCA. Six variables are selected (two internal pressures, one internal temperature, and three external temperatures) along with three additional variables related to sequence, cycle, and S10 fault logs. Furthermore, three statistical features — mean, median, and variance — are integrated into the dataset. Categorical variables are transformed into numerical formats to preserve sequential information, and all numerical variables are standardize. To reduce data imbalance, a particular emphasis on the data and knowledge from sequence S10 and the sequence immediately preceding it is made, enhancing the capture and anticipation of S10 faults. Following this process, 73 S10 faults are identified in a dataset of 70,000 observations, which are resampled at 15 min intervals, resulting in 4666 observations and a total of 12 variables for the next stages of analysis.

Scenario 2, incorporates a rule-based model into a knowledge base, leveraging a deep understanding of the equipment and its FMECA to enhance data preprocessing. The rules introduced not only improve the handling of missing values and outliers but also refine data reduction by validating the selection of the most relevant variables for predictive analysis. In Scenario 1, six variables are selected during the data reduction phase. However, in Scenario 2, this process is enhanced, resulting in the selection of eight variables through PCA, which are further validated by equipment monitoring rules alongside three key variables from sequence, cycle, and S10 fault logs. Additionally, the integration of further statistical features and FMECA insights allows for a more comprehensive characterization of the faults to be predicted.

This integration of knowledge has a significant impact, enabling the detection of additional faults that were not identified in the raw data from the automation system. In fact, Scenario 2 identifies 1490 S10 faults, a substantial increase compared to the 73 faults detected in Scenario 1. This clearly demonstrates that the inclusion of domain-specific knowledge captures more details and nuances in the data, leading to a more thorough fault detection process.

In terms of data volume, Scenario 2 produces a final dataset of 89,912 observations, resampled at 15 min intervals, resulting in 5994 observations, among which 1490 contain faults to be analyzed. The total number of variables is expanded to 18. This notable increase in the number of detected faults underscores the effectiveness of the knowledge-integrated approach, enabling more precise analysis and better preparation for predictive modeling.

The following Table 4 lists the final preprocessed data variables used in both scenarios: variables common to both scenarios are highlighted in gray, those specific to Scenario 1 in cyan, and those specific to Scenario 2 in white.

Below Fig. 8 provides descriptive statistics such as mean, median, and standard deviation for the continuous variables in the dataset. One can observe that the statistical indicators vary significantly from one cycle to another for the variables Var.1, Var.15, and Var.18, while they remain relatively constant for the other variables.

### 4.5.2. Sampling faults prediction results

This study evaluates the performance of three ML models — Random Forest, LightGBM and SVM classifier — across different prediction horizons 3 h, 12 h, and 24 h in two scenarios. The results, summarized in Fig. 9 and supported by Table 5, demonstrate significant differences in model performance between the two scenarios, with Scenario 2 clearly outperforming Scenario 1.

In Scenario 1, LightGBM emerges as the most effective model for short-term predictions, achieving an accuracy of 99.32% and an F1 score of 56.36% at the 3-h horizon. However, its performance declines sharply as the prediction horizon extends. By 12 h, the F1 score drops to 32.21%, and it further decreases to 12.43% at 24 h, highlighting a significant reduction in the model's predictive robustness over longer periods. Similarly, Random Forest and SVM classifier demonstrate strong performance at the 3-h horizon, although their results are slightly less impressive than LightGBM in this short-term window. Both models struggle to maintain their effectiveness beyond this point. Their F1 scores at 12 and 24 h indicate substantial difficulties in balancing precision and recall, which significantly limits their utility for longer prediction horizons within this scenario.

In Scenario 2, the integration of domain-specific knowledge into the predictive models leads to substantial and consistent improvements,

**Table 4**

Final preprocessed dataset used in both scenarios: in gray the ones common to both scenarios, in cyan the ones specific to scenario 1 and in white the ones specific to scenario 2.

| Variable | Description | Units |
|---|---|---|
| Var.1, Var.16 | Internal pressures | hPa |
| Var.18 | Internal temperature | °C |
| Var.7, Var.11, Var.12, Var.15 | External temperatures | °C |
| Var.17 | Angle measure | degree (°) |
| Mean | Mean for each observation | – |
| Median | Median for each observation | – |
| Variance | Variance for each observation | – |
| Sequence | Sequence variables | Discrete |
| Cycle number | Number of cycles (55 cycles in total) | Discrete |
| S10 Fault | S10 fault logs (S10 faults logs to be predicted in scenario 1 ) | – |
| Fault severity | Severity of the fault (1 if blocking and 0 if non-blocking) | Binary |
| Fault cause | Causes of fault appearance (high pressure, valve problems, etc.), each cause as a binary variable | Binary |
| Fault consequence | Consequence of fault appearance (1 if cycle stop and maintenance, 0 if fault acknowledgment) | Binary |
| Fault prioritization | Fault priority: 1 if part of the top 10 most important faults, 0 otherwise | Binary |
| Predictive target (Reconstructed) | Fault variable to be predicted | Binary |

**Table 5**

Performance metrics of combined ML Models for both scenarios and all horizons.

| Scenario | Model | Accuracy | | | | F1 Score | | | |
|---|---|---|---|---|---|---|---|---|---|
| | | 3 h | 12 h | 24 h | Avg | 3 h | 12 h | 24 h | Avg |
| Scenario 1 | Random Forest | 91.01% | 91.24% | 71.11% | 84.45% | 33.23% | 21.20% | 10.02% | 21.48% |
| | LightGBM | 99.32% | 98.40% | 74.33% | 90.68% | 56.36% | 32.21% | 12.43% | 33.67% |
| | SVM Classifier | 98.17% | 93.01% | 62.44% | 84.54% | 44% | 23.33% | 17.12% | 28.15% |
| Scenario 2 | Random Forest | 96.01% | 94% | 86.23% | 92.08% | 83.45% | 72.35% | 62.21% | 72.67% |
| | LightGBM | 99.15% | 98.03% | 99.21% | 98.80% | 90.36% | 91.46% | 93.12% | 91.65% |
| | SVM Classifier | 98.43% | 96.33% | 94.07% | 96.28% | 88% | 85.24% | 74% | 82.41% |

particularly for predictions beyond 3 hours—an area where Scenario 1 models notably failed. At 12 h, LightGBM achieves a remarkable accuracy of 98.03% and a robust F1 score of 91.46%, representing a dramatic enhancement compared to its performance in Scenario 1. This underscores the model's significantly improved capacity for balanced, long-term predictions when domain knowledge is incorporated. At 24 h, LightGBM continues to excel, maintaining an accuracy of 99.21% and an F1 score of 93.12%. This performance is in stark contrast to its Scenario 1 results and demonstrates the model's ability to reliably predict faults over extended horizons when improved with domain knowledge. Random Forest and SVM classifier also exhibit considerable improvements in Scenario 2. At the 24-h horizon, they achieve F1 scores of 62.21% and 74%, respectively, marking a significant enhancement over their Scenario 1 performance. Nonetheless, LightGBM consistently outperforms these models, particularly at longer prediction horizons, solidifying its position as the most reliable model within the knowledge-enhanced scenario.

Overall, the results from Scenario 1 indicate that the models are generally not effective beyond the 3-h horizon, highlighting their limitations in long-term predictive capability. In contrast, Scenario 2 demonstrates a remarkable improvement, enabling the models to extend their prediction reach up to 24 h with substantial reliability. This represents a significant advancement over Scenario 1, underscoring the critical role of domain knowledge in enhancing the performance and robustness of ML models for long-term predictions.

### 4.5.3. Comparison between the current state and Scenario 1 and 2

This part compares the results obtained in Scenario 1 and 2 with current monitoring model (Table 6), focusing on identifying the best predictive model capable of anticipating blocking faults over different time horizons with strong performance.

**Table 6**

Summary table: Current state of the process, scenario 1 and scenario 2 results.

| Comparative elements | PdM | | | |
|---|---|---|---|---|
| | | Baseline: condition monitoring | Scenario 1: Purely data driven | Scenario 2: enhanced hybrid approach |
| Best model | | Rule-based model | LightGBM | LightGBM |
| Best prediction horizon achieved | | 0 h | 3 h | 24 h |
| Performance achieved | | Basic condition-based monitoring, prone to faults | Accuracy: 99.32%, F1 score: 56.36% | Accuracy: 99.21% F1 score: 93.12% |

Currently, the studied equipment is monitored using thresholds of physical measurements provided by a rule-based model, ensuring the smooth progress of sequences until cycle completion. However, this model has limitations in terms of responsiveness, as it requires the occurrence of a fault to trigger actions. Therefore, the predictive horizon is zero, and the model performance is constrained by insufficient monthly sampling (6 samples taken per month instead of 15), primarily due to a high rate of sampling faults.

Scenario 1, employing a purely data-driven approach, has improved this situation by predicting sampling faults with better performance metrics (accuracy : 99.32%, F1 score: 56.36%). Despite a predictive horizon limited to 3 h and a moderate F1 score (indicating a moderate precision of the LightGBM model), this approach represents significant

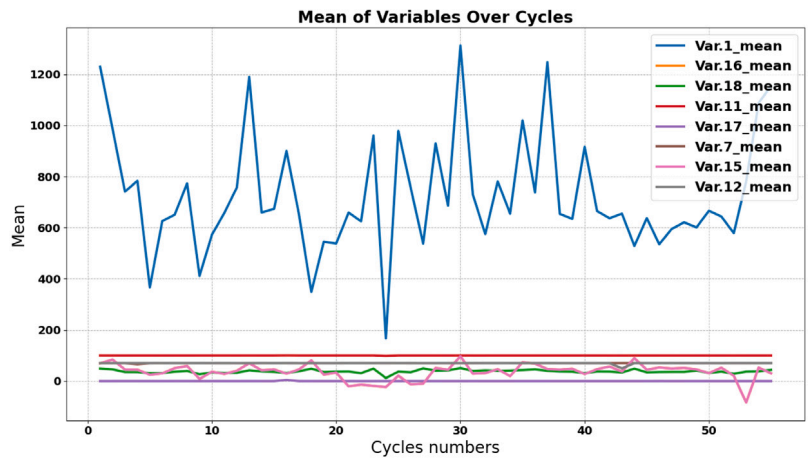

(a) Evolution of the mean of variables over update cycles.

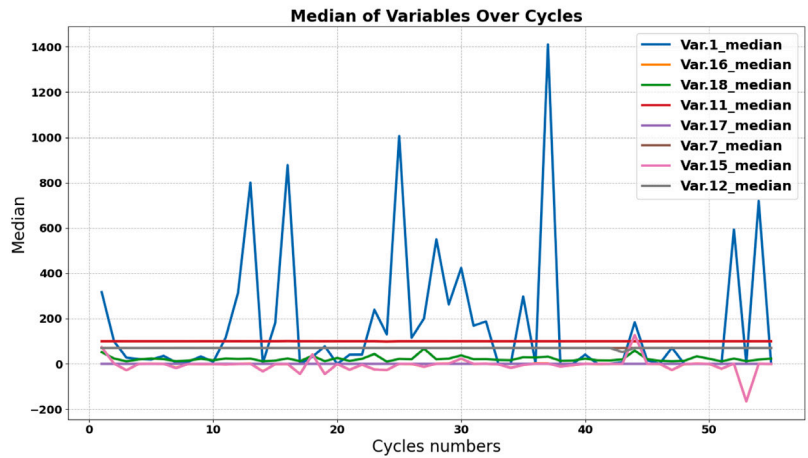

(b) Evolution of the median of variables over update cycles.

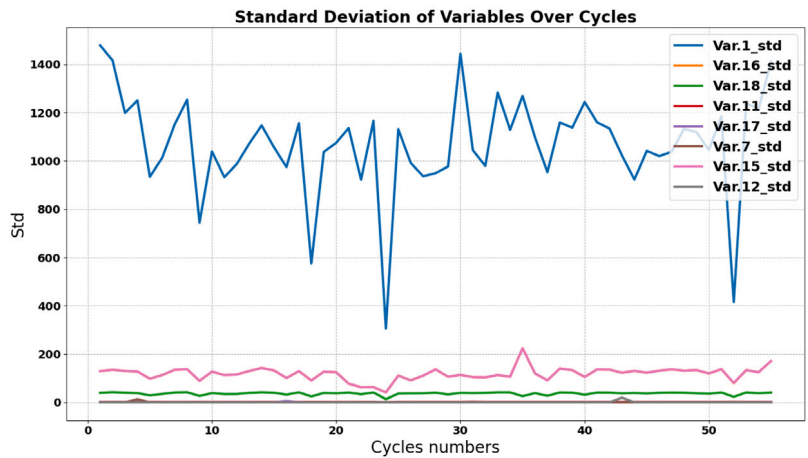

(c) Evolution of the Standard Deviation of variables over update cycles.

**Fig. 8.** Comparative analysis of various statistics over cycles.

progress compared to the current model, although the timeframe is short for effective preventive maintenance planning.

By integrating the current model knowledge with scenario 1 through scenario 2, better results are achieved with the same model: LightGBM. This scenario allows prediction of sampling faults 24 h before sequence start, with improved performance (accuracy : 99.21%, F1 score : 93.12%). The hybridization between the current rule-based method used to monitor the process and the data-driven approach in Scenario 1 highlights the importance of combining both data and knowledge to improve the reliability and effectiveness of PdM. In conclusion, these results demonstrate a significant improvement in the responsiveness and performance of PdM, transitioning from a rule-based model to more advanced data-driven approaches, or a combination thereof. This opens the door to more proactive and efficient interventions to optimize industrial equipment availability.

### 4.6. Discussion

The comparison of Scenarios 1 and 2 with the current state of PdM reveals important insights into the effectiveness of different approaches.

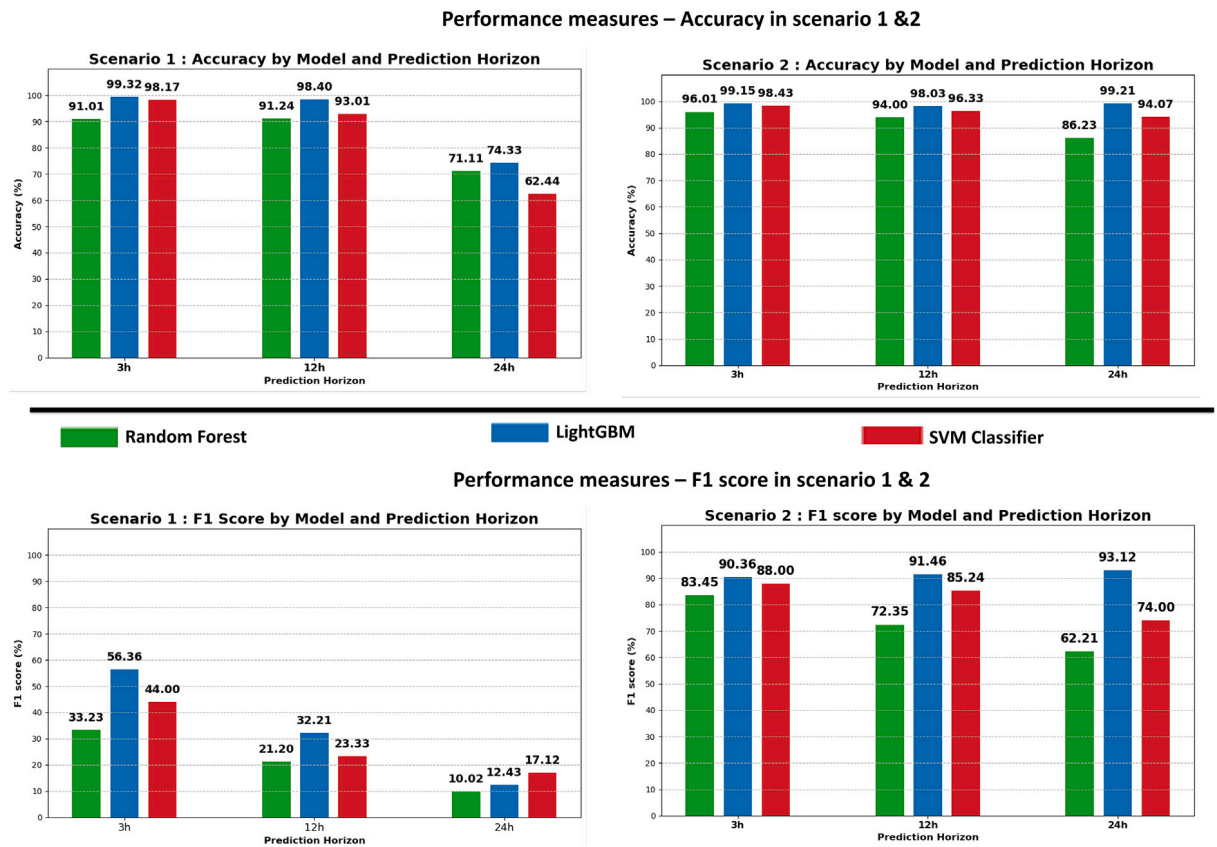

**Fig. 9.** Comparison of Accuracy and F1 Scores across Scenarios 1 and 2 with Different Prediction Horizons: 3 h, 12 h and 24 h. The top panels of the figure show the accuracy of the models while the bottom panels illustrate the corresponding F1 scores.

Firstly, Scenario 1, utilizing a purely data-driven technique, demonstrates notable improvements over the current rule-based monitoring system. Despite its predictive horizon being limited to 3 h and achieving a moderate F1 score (56.36%), Scenario 1 showcases enhanced capabilities in predicting equipment faults. This shift towards data-driven methods highlights the potential for increased responsiveness and efficiency in maintenance operations.

Secondly, Scenario 2, which combines both rule-based and data-driven approaches, presents even more promising results. By integrating the current rule-based monitoring system with data-driven predictive models, Scenario 2 achieves a predictive horizon of 24 h and significantly enhances performance metrics, with an accuracy of 99.21% and an F1 score of 93.12%. This hybrid approach leverages the strengths of both methodologies, resulting in improved reliability and predictive precision. The optimal predictive horizon achieved in Scenario 2 allows for more proactive maintenance planning, enabling organizations to address potential equipment issues before they occur. Additionally, the higher accuracy and F1 score mean better precision and effectiveness in fault detection and prediction, leading to reduced downtime and maintenance costs.

Overall, the findings emphasize the importance of integrating data-driven techniques with existing rule-based systems in PdM. The successful validation of the proposed methodology within Scenario 2 confirms the value of this hybrid approach. By combining these approaches, organizations can leverage historical data and domain knowledge to enhance maintenance strategies, ultimately improving equipment reliability and operational efficiency.

The methodology demonstrated ability to achieve a 24-h predictive horizon on the studied case study and significantly improve performance metrics highlights the tangible benefits of this integration. In summary, the validation of the methodology within Scenario 2 underscores the relevance and effectiveness of integrating data-driven methodologies into existing maintenance practices. This combined approach paves the way for continuous improvements in asset management and contributes to maximizing the value of technological investments in PdM.

## 5. Conclusion and future work

In the context of Industry 4.0, the integration of technologies such as IoT and Artificial Intelligence has encouraged the development of data-driven approaches for improving the availability and reliability of industrial systems. The increasing availability of sensor data and the development of advanced analytics have strengthened the role of PHM in supporting PdM strategies. While ML-based methods are increasingly used in PdM, they often suffer from limitations such as a lack of interpretability and the need for extensive domain knowledge to guide the models. Moreover, these methods assume that the available data fully reflect the real monitoring conditions of the equipment—an assumption that rarely holds in practice. These challenges are even more critical in the nuclear sector, where systems are highly complex, safety requirements are stringent, and effective monitoring often requires close integration with domain knowledge. Thus, in such contexts, the effectiveness of data-driven methodologies depends heavily on the incorporation of knowledge to ensure meaningful outcomes. In many existing works, the hybridization of data and domain knowledge is handled in an ad hoc manner, often limited to the decision stage, thus reducing its contribution to model performance and robustness.

This paper addresses these challenges by proposing a hybrid methodology for PdM that integrates structured domain knowledge into data-driven ML models. The knowledge, drawn from sources such as FMECA analyses, expert interviews, and maintenance history, is formalized and integrated into two key stages of the modeling process. First, during the data preprocessing phase, the structured knowledge help improve data quality by guiding outlier detection, correcting false anomalies, and enriching features with relevant knowledge. Second, during model selection, they assist in choosing algorithms that best reflect the nature of the targeted failures.

The proposed methodology is validated through a real-world case study in the nuclear industry, comparing different scenarios with the current state of PdM. Results show that incorporating domain knowledge improves predictive performance while significantly extending the prediction horizon. In particular, Scenario 2 (applying the hybrid approach) achieved a 24-h prediction horizon with an accuracy of 99.21% and an F1 score of 93.12%, outperforming both purely data-driven approach in Scenario 1 and the existing rule-based monitoring system.

Future work will focus on deploying the proposed methodology in an online environment (Stage 4), investigating other knowledge integration techniques beyond rule-based systems (such as case-based reasoning or fuzzy logic), and testing the approach across additional case studies to assess its generalizability. Although the methodology was validated in a nuclear context, its structure can be adapted to other industrial domains requiring reliable and knowledge-informed predictive maintenance strategies. Further exploration of uncertainty management techniques could also help enhance the robustness and trustworthiness of predictive models in critical industrial applications.

## CRediT authorship contribution statement

**Amaratou Mahamadou Saley:** Writing – original draft, Data curation, Methodology, Software, Writing – review & editing, Formal analysis. **Thierry Moyaux:** Supervision, Writing – review & editing. **Aïcha Sekhari:** Supervision. **Vincent Cheutet:** Supervision, Validation, Writing – review & editing. **Jean-Baptiste Danielou:** Supervision.

## Acknowledgments

For the purpose of Open Access, a CC-BY public copyright licence has been applied by the authors to the present document and will be applied to all subsequent versions up to the Author Accepted Manuscript arising from this submission.

## Data availability

The data that has been used is confidential.

## References

Abdelillah, F. M., Nora, H., Samir, O., & Sidi-Mohammed, B. (2023). Hybrid data-driven and knowledge-based predictive maintenance framework in the context of industry 4.0. In *International conference on model and data engineering* (pp. 319–337). Sousse, Tunisia: Springer, http://dx.doi.org/10.1007/978-3-031-49333-1_23.

Aboshosha, A., Haggag, A., George, N., & Hamad, H. A. (2023). IoT-based data-driven predictive maintenance relying on fuzzy system and artificial neural networks. *Scientific Reports*, *13*(1), 12186. http://dx.doi.org/10.1038/s41598-023-38887-z.

Aizpurua, J. I., McArthur, S. D. J., Stewart, B. G., Lambert, B., Cross, J. G., & Catterson, V. M. (2019). Adaptive power transformer lifetime predictions through machine learning and uncertainty modeling in nuclear power plants. *IEEE Transactions on Industrial Electronics*, *66*(6), 4726–4737. http://dx.doi.org/10.1109/TIE.2018.2860532.

Al-Refaie, A., Al-atrash, M., & Lepkova, N. (2025). Prediction of the remaining useful life of a milling machine using machine learning. *MethodsX*, *14*, Article 103195. http://dx.doi.org/10.1016/j.mex.2025.103195, URL https://www.sciencedirect.com/science/article/pii/S2215016125000433.

Allal, Z., Noura, H. N., Vernier, F., Salman, O., & Chahine, K. (2024). Wind turbine fault detection and identification using a two-tier machine learning framework. *Intelligent Systems with Applications*, *22*, Article 200372. http://dx.doi.org/10.1016/j.iswa.2024.200372, URL https://www.sciencedirect.com/science/article/pii/S2667305324000486.

Alshboul, O., Al Mamlook, R. E., Shehadeh, A., & Munir, T. (2024). Empirical exploration of predictive maintenance in concrete manufacturing: Harnessing machine learning for enhanced equipment reliability in construction project management. *Computers & Industrial Engineering*, *190*, Article 110046. http://dx.doi.org/10.1016/j.cie.2024.110046, URL https://www.sciencedirect.com/science/article/pii/S0360835224001670.

Ayadi, A. (2018). *semantic approaches for the meta-optimization of complex biomolecular networks* (Ph.D. thesis), Université de Strasbourg ; Institut supérieur de gestion (Tunis).

Barry, I., & Hafsi, M. (2023). Towards hybrid predictive maintenance for aircraft engine: Embracing an ontological-data approach. In *2023 ACS/IEEE international conference on computer systems and applications* (pp. 1–6). Giza, Egypt: http://dx.doi.org/10.1109/AICCSA59173.2023.10479328.

Bouhadra, K., & Forest, F. (2024). Knowledge-based and expert systems in prognostics and health management: a survey. *International Journal of Prognostics and Health Management*, *15*(2), http://dx.doi.org/10.36001/ijphm.2024.v15i2.3986.

Cai, C., Jiang, Z., Wu, H., Wang, J., Liu, J., & Song, L. (2024). Research on knowledge graph-driven equipment fault diagnosis method for intelligent manufacturing. *International Journal of Advanced Manufacturing Technology*, *130*(9), 4649–4662. http://dx.doi.org/10.1007/s00170-024-12998-x.

Cancemi, S., Angelucci, M., Chierici, A., Paci, S., & Frano, R. L. (2025). Hybrid neural network and statistical forecasting methodology for predictive monitoring and residual useful life estimation in nuclear power plant components. *Nuclear Engineering and Design*, *433*, Article 113900. http://dx.doi.org/10.1016/j.nucengdes.2025.113900, URL https://www.sciencedirect.com/science/article/pii/S0029549325000779.

Cao, Q., Zanni-Merk, C., Samet, A., De Beuvron, F. D. B., & Reich, C. (2020). Using rule quality measures for rule base refinement in knowledge-based predictive maintenance systems. *Cybernetics and Systems*, *51*(2), 161–176. http://dx.doi.org/10.1080/01969722.2019.1705550.

Carvalho, T. P., Soares, F. A., Vita, R., Francisco, R. d. P., Basto, J. P., & Alcalá, S. G. (2019). A systematic literature review of machine learning methods applied to predictive maintenance. *Computers & Industrial Engineering*, *137*, Article 106024. http://dx.doi.org/10.1016/j.cie.2019.106024.

Chapelin, J., Voisin, A., Rose, B., t Iung, B., Steck, L., Chaves, L., Lauer, M., & Jotz, O. (2025). Data-driven drift detection and diagnosis framework for predictive maintenance of heterogeneous production processes: Application to a multiple tapping process. *Engineering Applications of Artificial Intelligence*, *139*, Article 109552. http://dx.doi.org/10.1016/j.engappai.2024.109552, URL https://www.sciencedirect.com/science/article/pii/S095219762401710X.

Chen, C., Shi, J., Shen, M., Feng, L., & Tao, G. (2023). A predictive maintenance strategy using deep learning quantile regression and kernel density estimation for failure prediction. *IEEE Transactions on Instrumentation and Measurement*, *72*, http://dx.doi.org/10.1109/TIM.2023.3240208.

Chiang, L. H., Russell, E. L., & Braatz, R. D. (2001). Knowledge-based methods. In *Fault detection and diagnosis in industrial systems* (pp. 223–254). London: Springer.

Cummins, L., Sommers, A., Ramezani, S. B., Mittal, S., Jabour, J., Seale, M., & Rahimi, S. (2024). Explainable predictive maintenance: A survey of current methods, challenges and opportunities. *IEEE Access*, *12*, 57574–57602. http://dx.doi.org/10.1109/ACCESS.2024.3391130.

Deepika, J., Reddy, P. M., Murari, K., & Rahul, B. (2025). Predictive maintenance of aircraft engines: Machine learning approaches for remaining useful life estimation. In *2025 6th international conference on mobile computing and sustainable informatics* (pp. 1679–1686). Goathgaun, Nepal: http://dx.doi.org/10.1109/ICMCSI64620.2025.10883526.

Diversi, R., Lenzi, A., Speciale, N., & Barbieri, M. (2025). An autoregressive-based motor current signature analysis approach for fault diagnosis of electric motor-driven mechanisms. *Sensors*, *25*(4), http://dx.doi.org/10.3390/s25041130, URL https://www.mdpi.com/1424-8220/25/4/1130.

Dong, E., Zhan, X., Yan, H., Tan, S., Bai, Y., Wang, R., & Cheng, Z. (2025). A data-driven intelligent predictive maintenance decision framework for mechanical systems integrating transformer and kernel density estimation. *Computers & Industrial Engineering*, *201*, Article 110868. http://dx.doi.org/10.1016/j.cie.2025.110868, URL https://www.sciencedirect.com/science/article/pii/S0360835225000130.

Ekpenyong, M. E., & Udoh, N. S. (2024). Intelligent optimal preventive replacement maintenance policy for non-repairable systems. *Computers & Industrial Engineering*, *190*, Article 110091. http://dx.doi.org/10.1016/j.cie.2024.110091, URL https://www.sciencedirect.com/science/article/pii/S0360835224002122.

Elkateb, S., Métwalli, A., Shendy, A., & Abu-Elanien, A. E. (2024). Machine learning and IoT – based predictive maintenance approach for industrial applications. *Alexandria Engineering Journal*, *88*, 298–309. http://dx.doi.org/10.1016/j.aej.2023.12.065, URL https://www.sciencedirect.com/science/article/pii/S1110016823011572.

Facchinetti, T., Arazzi, M., & Nocera, A. (2022). Time series forecasting for predictive maintenance of refrigeration systems. In *2022 IEEE international symposium on dependable, autonomic and secure computing* (pp. 378–383). Falerna, Italy: http://dx.doi.org/10.1109/DASC/PiCom/CBDCom/Cy55231.2022.9927978.

Ferrisi, S., Cappellari, P., Guido, R., Umbrello, D., & Ambrogio, G. (2025). Application of two-parameter Weibull distribution for predictive maintenance: A case study. In *6th International Conference on Industry 4.0 and Smart Manufacturing. Procedia Computer Science*, *253*, 3160–3168. http://dx.doi.org/10.1016/j.procs.2025.02.041.

Ganga, D., & Ramachandran, V. (2020). Adaptive prediction model for effective electrical machine maintenance. *Journal of Quality in Maintenance Engineering*, *26*(1), 166–180. http://dx.doi.org/10.1108/JQME-12-2017-0087.

Gawde, S., Patil, S., Kumar, S., Kamat, P., Kotecha, K., & Alfarhood, S. (2024). Explainable predictive maintenance of rotating machines using LIME, SHAP, PDP, ICE. *IEEE Access*, *12*, 29345–29361. http://dx.doi.org/10.1109/ACCESS.2024.3367110.

Gay, A. (2023). *Pronostic de défaillance basé sur les données pour la prise de décision en maintenance: Exploitation du principe d'augmentation de données avec intégration de connaissances à priori pour faire face aux problématiques du small data set (Ph.D. thesis)*, Université de Lorraine.

Gharib, H., & Kovács, G. (2024). Implementation and possibilities of fuzzy logic for optimal operation and maintenance of marine diesel engines. *Machines*, *12*(6), 425. http://dx.doi.org/10.3390/machines12060425.

Haïk, P., Mahé, S., & Ricard, B. (2002). Knowledge engineering as a support for decision making in plant operation and maintenance. In *Pacific rim knowledge acquition workshop*. Tokyo, Japan.

Huang, C., Bu, S., Lee, H. H., Chan, C. H., Kong, S. W., & Yung, W. K. (2024). Prognostics and health management for predictive maintenance: A review. *Journal of Manufacturing Systems*, *75*, 78–101. http://dx.doi.org/10.1016/j.jmsy.2024.05.021, URL https://www.sciencedirect.com/science/article/pii/S0278612524001183.

Huang, J. W., & Gao, J. W. (2020). How could data integrate with control? A review on data-based control strategy. *International Journal of Dynamics and Control*, *8*(4), 1189–1199. http://dx.doi.org/10.1007/s40435-020-00688-x.

ISO 13372 (2012). *Surveillance et diagnostic de l'état des machines- Vocabulaire*. ISO (Organisation internationale de normalisation).

ISO 13374-2 (2007). *Surveillance et diagnostics d'état des machines - Traitement, échange et présentation des données*. ISO (Organisation internationale de normalisation).

ISO 14224 (2017). *Collection and exchange of reliability and maintenance data for equipment*. ISO (Organisation internationale de normalisation).

ISO13379 (2015). *Condition monitoring and diagnostics of machines — Data interpretation and diagnostics techniques — Part 2: Data-driven applications*. ISO (Organisation internationale de normalisation).

Kapuria, A., & Cole, D. G. (2023). Integrating survival analysis with Bayesian statistics to forecast the remaining useful life of a centrifugal pump conditional to multiple fault types. *Energies*, *16*(9), http://dx.doi.org/10.3390/en16093707.

Kasilingam, S., Yang, R., Singh, S. K., Farahani, M. A., Rai, R., & Wuest, T. (2024). Physics-based and data-driven hybrid modeling in manufacturing: a review. *Production & Manufacturing Research*, *12*(1), Article 2305358. http://dx.doi.org/10.1080/21693277.2024.2305358.

Katerina, I., Alexandros, B., Apostolou, D., & Gregoris, M. (2020). Prescriptive analytics: Literature review and research challenges. *International Journal of Information Management*, *50*, 57–70. http://dx.doi.org/10.1016/j.ijinfomgt.2019.04.003.

Kızito, R., Scruggs, P., Li, X., Devinney, M., Jansen, J., & Kress, R. (2021). Long short-term memory networks for facility infrastructure failure and remaining useful life prediction. *IEEE Access*, *9*, 67585–67594. http://dx.doi.org/10.1109/ACCESS.2021.3077192.

Klein, P. (2025). *Combining Expert Knowledge and Deep Learning with Case-Based Reasoning for Predictive Maintenance*. Springer Nature.

Koksal, E. S., Asrav, T., Esenboga, E. E., Cosgun, A., Kusoglu, G., & Aydin, E. (2024). Physics-informed and data-driven modeling of an industrial wastewater treatment plant with actual validation. *Computers & Chemical Engineering*, *189*, Article 108801. http://dx.doi.org/10.1016/j.compchemeng.2024.108801.

Li, Z., Jana, C., Pamucar, D., & Pedrycz, W. (2025). A comprehensive assessment of machine learning models for predictive maintenance using a decision-making framework in the industrial sector. *Alexandria Engineering Journal*, *120*, 561–583. http://dx.doi.org/10.1016/j.aej.2025.02.010.

Li, C., Li, S., Feng, Y., Gryllias, K., Gu, F., & Pecht, M. (2024). Small data challenges for intelligent prognostics and health management: a review. *Artificial Intelligence Review*, *57*(8), 214. http://dx.doi.org/10.1007/s10462-024-10820-4.

Liu, X., Cheng, W., Xing, J., Chen, X., Li, L., Guan, Y., Ding, B., Nie, Z., Zhang, R., & Zhi, Y. (2024). Predictive maintenance system for high-end equipment in nuclear power plant under limited degradation knowledge. *Advanced Engineering Informatics*, *61*, Article 102506.

Luk, S. S., Jin, Y., Zhang, X., Ng, V. T. Y., Huang, J., & Wong, C. N. (2025). Condition assessment and predictive maintenance for contact probe using health index and encoder-decoder LSTM model. *Quality and Reliability Engineering International*, *41*(1), 154–173. http://dx.doi.org/10.1002/qre.3668.

Ma, J., & Jiang, J. (2011). Applications of fault detection and diagnosis methods in nuclear power plants: A review. *Progress in Nuclear Energy*, *53*(3), 255–266. http://dx.doi.org/10.1016/j.pnucene.2010.12.001.

Mokhtarzadeh, M., Rodríguez-Echeverría, J., Semanjski, I., & Gautama, S. (2024). Hybrid intelligence failure analysis for industry 4.0: a literature review and future prospective. *Journal of Intelligent Manufacturing*, 1–26. http://dx.doi.org/10.1007/s10845-024-02376-5.

Mrad, S., & Mraihi, R. (2023). An overview of model-driven and data-driven forecasting methods for smart transportation. In *Data analytics and computational intelligence: novel models, algorithms and applications* (pp. 159–183). Cham: Springer Nature Switzerland, http://dx.doi.org/10.1007/978-3-031-38325-0_8.

Nemeth, T., Ansari, F., Sihn, W., Haslhofer, B., & Schindler, A. (2018). PriMa-X: A reference model for realizing prescriptive maintenance and assessing its maturity enhanced by machine learning. *Procedia CIRP*, *72*(June), 1039–1044. http://dx.doi.org/10.1016/j.procir.2018.03.280.

Nie, J., Jiang, J., Li, Y., Wang, H., Ercisli, S., & Lv, C. (2025). Data and domain knowledge dual-driven artificial intelligence: Survey, applications, and challenges. *Expert Systems*, *42*(1), Article e13425. http://dx.doi.org/10.1111/exsy.13425.

Nunes, P., Santos, J., & Rocha, E. (2023). Challenges in predictive maintenance – a review. *CIRP Journal of Manufacturing Science and Technology*, *40*, 53–67. http://dx.doi.org/10.1016/j.cirpj.2022.11.004.

Ozgür-Unlüakın, D., & Bilgiç, T. (2006). Predictive maintenance using dynamic probabilistic networks. In *Proceedings of the 3rd European workshop on probabilistic graphical models pGM'06* (pp. 141–148). Prague, Czech Republic.

Peng, Y., Dong, M., & Zuo, M. J. (2010). Current status of machine prognostics in condition-based maintenance: A review. *International Journal of Advanced Manufacturing Technology*, *50*(1–4), 297–313. http://dx.doi.org/10.1007/s00170-009-2482-0.

Pierleoni, P., Palma, L., Belli, A., Raggiunto, S., & Sabbatini, L. (2022). Supervised regression learning for maintenance-related data. In G. Fortino, R. Gravina, A. Guerrieri, & C. Savaglio (Eds.), *2022 IEEE international symposium on dependable, autonomic and secure computing* (pp. 360–365). Falerna, Italy: http://dx.doi.org/10.1109/DASC/PiCom/CBDCom/Cy55231.2022.9927904.

Rajaoarisoa, L., Randrianandraina, R., Nalepa, G. J., & Gama, J. (2025). Decision-making systems improvement based on explainable artificial intelligence approaches for predictive maintenance. *Engineering Applications of Artificial Intelligence*, *139*, Article 109601. http://dx.doi.org/10.1016/j.engappai.2024.109601, URL https://www.sciencedirect.com/science/article/pii/S0952197624017597.

Sadeghi, J., & Askarinejad, H. (2010). Development of improved railway track degradation models. *Structure and Infrastructure Engineering*, *6*(6), 675–688. http://dx.doi.org/10.1080/15732470801902436.

Saley, A. M., Marchand, J., Sekhari, A., Cheutet, V., & Danielou, J. B. (2022). State-of-art and maturity overview of the nuclear industry on predictive maintenance. In *IFIP international conference on product lifecycle management* (pp. 337–346). Grenoble, France: Springer, http://dx.doi.org/10.1007/978-3-031-25182-5_33.

Shukla, K., Nefti-Meziani, S., & Davis, S. (2022). A heuristic approach on predictive maintenance techniques: Limitations and scope. *Advances in Mechanical Engineering*, *14*(6), Article 168781322211010009. http://dx.doi.org/10.1177/16878132221101009.

Si, X. S., Wang, W., Hu, C. H., & Zhou, D. H. (2011). Remaining useful life estimation - A review on the statistical data driven approaches. *European Journal of Operational Research*, *213*(1), 1–14. http://dx.doi.org/10.1016/j.ejor.2010.11.018.

Suebsombut, P. (2021). *adaptive decision support system for smart agricultural crop cultivation to support food safety standard (Ph.D. thesis)*, Université Lumière Lyon2 (France) & Chiang Mai University (Thailand).

Susto, G. A., Schirru, A., Pampuri, S., McLoone, S., & Beghi, A. (2014). Machine learning for predictive maintenance: A multiple classifier approach. *IEEE Transactions on Industrial Informatics*, *11*(3), 812–820. http://dx.doi.org/10.1109/TII.2014.2349359.

Tran, V. T., Thom Pham, H., Yang, B. S., & Tien Nguyen, T. (2012). Machine performance degradation assessment and remaining useful life prediction using proportional hazard model and support vector machine. *Mechanical Systems and Signal Processing*, *32*, 320–330. http://dx.doi.org/10.1016/j.ymssp.2012.02.015, URL https://www.sciencedirect.com/science/article/pii/S0888327012000512. Uncertainties in Structural Dynamics.

Ubesigha, L., Rajapaksha, C., Mahanama, T., & Asanka, D. (2025). Enhancing predictive maintenance in industry 4.0: Leveraging AI for optimal machinery health management. In *2025 5th international conference on advanced research in computing* (pp. 1–6). http://dx.doi.org/10.1109/ICARC64760.2025.10963100.

Vrignat, P., Avila, M., Duculty, F., & Kratz, F. (2015). Failure event prediction using hidden Markov model approaches. *IEEE Transactions on Reliability*, *64*(3), 1038–1048. http://dx.doi.org/10.1109/TR.2015.2423191.

Wang, Y., Shen, F., & Ye, L. (2025). A knowledge-refined hybrid graph model for quality prediction of industrial processes. *Engineering Applications of Artificial Intelligence*, *139*, Article 109711. http://dx.doi.org/10.1016/j.engappai.2024.109711, URL https://www.sciencedirect.com/science/article/pii/S0952197624018694.

Wei, F., Tan, L., Ma, X., Xiao, H., Patel, D., Lee, C. G., & Yang, L. (2025). A hybrid prognostic framework: Stochastic degradation process with adaptive trajectory learning to transfer historical health knowledge. *Mechanical Systems and Signal Processing*, *224*, Article 112171. http://dx.doi.org/10.1016/j.ymssp.2024.112171.

Xu, A., Wang, R., Weng, X., Wu, Q., & Zhuang, L. (2025). Strategic integration of adaptive sampling and ensemble techniques in federated learning for aircraft engine remaining useful life prediction. *Applied Soft Computing*, *175*, Article 113067. http://dx.doi.org/10.1016/j.asoc.2025.113067, URL https://www.sciencedirect.com/science/article/pii/S1568494625003783.

Yan, R., Zhou, Z., Shang, Z., Wang, Z., Hu, C., Li, Y., Yang, Y., Chen, X., & Gao, R. X. (2025). Knowledge driven machine learning towards interpretable intelligent prognostics and health management: Review and case study. *Chinese Journal of Mechanical Engineering*, *38*(1), 5. http://dx.doi.org/10.1186/s10033-024-01173-8.

Yang, L., & Liao, Y. (2024). An optimal data-driven framework for RUL prediction with uncertainty quantification. *IEEE Sensors Journal*, http://dx.doi.org/10.1109/JSEN.2024.3510720.

Zanotelli, M., Hines, J. W., & Coble, J. B. (2024). Combining similarity measures and left-right hidden Markov models for prognostics of items subjected to perfect and imperfect maintenance. *Nuclear Science and Engineering*, http://dx.doi.org/10.1080/00295639.2024.2303165.

Zenisek, J., Holzinger, F., & Affenzeller, M. (2019). Machine learning based concept drift detection for predictive maintenance. *Computers & Industrial Engineering, 137*, Article 106031. http://dx.doi.org/10.1016/j.cie.2019.106031.

Zhang, X., Li, L., Zhang, X., Song, Z., & Qian, J. (2024). Gaussian–Poisson mixture regression model for defects prediction in steelmaking. *Chemometrics and Intelligent Laboratory Systems, 246*, Article 105088. http://dx.doi.org/10.1016/j.chemolab.2024.105088.

Zhuang, J., Jia, M., Huang, C. G., Beer, M., & Feng, K. (2024). Health prognosis of bearings based on transferable autoregressive recurrent adaptation with few-shot learning. *Mechanical Systems and Signal Processing, 211*, Article 111186. http://dx.doi.org/10.1016/j.ymssp.2024.111186.